\documentclass{bmvc2k}

\usepackage{graphicx}
\usepackage{amsmath}
\usepackage{amssymb}
\usepackage{booktabs}

\usepackage{pifont}% http://ctan.org/pkg/pifont
\usepackage{epsfig}
\usepackage{graphicx}
\usepackage{amsmath}
\usepackage{amssymb}
\usepackage{float}
\usepackage{epsfig}
\usepackage{booktabs}
\usepackage{multirow}
\usepackage[ruled,vlined]{algorithm2e}
\usepackage{multirow}
\usepackage{comment}

\newcommand{\yasu}[1]{\textcolor{orange}{\{yasu: {#1}\}}}

\title{Floorplan Restoration by Structure Hallucinating Transformer Cascades}

\addauthor{Sepidehsadat Hosseini}{
sepidh@sfu.ca
}{1}
\addauthor{Yasutaka Furukawa}{furukawa@sfu.ca}{1}

\addinstitution{
GrUVi Lab, VML Lab\\
Simon Fraser University\\
 British Colombia, BC
}

\usepackage{expl3}
\ExplSyntaxOn
\newcommand\latinabbrev[1]{
  \peek_meaning:NTF . {% Same as \@ifnextchar
    #1\@}%
  { \peek_catcode:NTF a {% Check whether next char has same catcode as \'a, i.e., is a letter
      #1.\@ }%
    {#1.\@}}}
\ExplSyntaxOff

\def\eg{\latinabbrev{e.g}}
\def\etal{\latinabbrev{et al}}

\def\ie{\latinabbrev{i.e}}
\newcommand{\mysubsubsection}[1]{\vspace{0.1cm} \noindent {\bf #1}:}

\runninghead{Hosseini, Furukawa}{Floorplan Restoration}

\def\eg{\emph{e.g}\bmvaOneDot}

\def\etal{\emph{et al}\bmvaOneDot}

\begin{document}

\maketitle

    %\centerline{
    %\includegraphics[width=\textwidth]{figures/tea.pdf} }
    %\captionof{figure}{The paper presents a new extreme floorplan %reconstruction task. An input is a partial floorplan %inferred or curated from panorama images of a %house/apartment. An output is a complete floorplan. A %technical challenge is the hallucination of invisible rooms %and doors. %The paper presents a new benchmark and a neural %architecture for a solution.}
    %}
    %\label{fig:my_label}

%\vspace{1em}

\begin{abstract}

This paper presents a novel floorplan restoration
task, a new benchmark for the task, and a neural architecture as a solution. Given a partial floorplan reconstruction inferred from panorama images, the task is to restore a complete floorplan including invisible architectural structures. The proposed neural network 1)
encodes an input partial floorplan into a set of latent vectors by convolutional neural networks and a Transformer;
and 2) recovers an entire floorplan while hallucinating invisible rooms and doors by cascading Transformer
decoders. Qualitative and quantitative evaluations demonstrate the effectiveness of our approach over the benchmark of
701 houses, outperforming the state-of-the-art reconstruction techniques. We will publish code and data.
\end{abstract}

\begin{comment}This paper presents an extreme floorplan reconstruction task, a new benchmark for the task, and a neural architecture as a solution. 
Given a partial floorplan reconstruction inferred or curated from panorama images,
%posed panorama images of a residential indoor space (e.g., from a real-estate listing), 
the task is to reconstruct a complete floorplan including invisible architectural structures. The proposed neural network 1) encodes an input partial floorplan into a set of latent vectors by convolutional neural networks and a Transformer; and 2) reconstructs an entire floorplan while hallucinating invisible rooms and doors by cascading Transformer decoders.
%a visible room/door encoder, which uses convolutional neural networks and a Transformer to and invisible room/door decoder consisting of cascading 
%Transformers component-wise transformers encoding panorama images into a set of latent vectors and 2) cascading transformer decoders reconstructing an entire floorplan including invisible rooms and doors.
%restoreing invisible rooms and doors in three stages.
%forms a cascade of transformer decoders to hallucinate invisible rooms and doors.
%
%The paper provides a new benchmark of 701 houses with ground-truth vector-graphics floorplans.
%2,355 panorama images. 
Qualitative and quantitative evaluations demonstrate effectiveness of our approach over the benchmark of 701 houses, outperforming the state-of-the-art reconstruction techniques. We will share our code, models, and data.
\end{comment}

\section{Introduction}
%Either architecture has regularities or panorama indoor scanning explosion. Probably explosion.

Indoor panorama photography is exploding. Pioneered by Ricoh Theta~\cite{ricoh}, consumer-grade panorama cameras are prevalent on the market, whose applications range from real estate to entertainment and surveillance.  Ricoh Theta cameras have collected 100 million panoramas from residential houses, which
%. This indoor panorama collection allows
allow house renters/buyers/realtors to browse immersive 360 views for tens of millions of houses. However, these panorama collections are extremely sparse (\ie, one panorama per room with little visual overlaps) and even leave some rooms invisible, posing fundamental challenges to existing techniques.
%to enable more advanced applications.

This paper presents a new floorplan restoration task, a benchmark, and a solution, which could exploit 100 million panoramas to create floorplans for tens of millions of houses. %for more sophisticated architectural analysis.
The potential applications range from real estate and construction industries (\eg, building code verification and property value assessment) to virtual and augmented reality.
%(e.g., digital content creation).
%
Concretely, the task is to take
%focuses on the task of floorplan reconstruction, where an input is 
a partial floorplan inferred or manually curated from panorama images, and restore a complete floorplan.
Note that the task is named ``restoration'', which is too ambiguous as reconstruction, but has a unique solution, and is more constrained than inpainting.
%an output is a complete floorplan.
%{\color{red} Here better explain why we don't do generation but reconstruction, also telling how Reconstruction is possible as architectures follow general rules to draw floorplan and our network proves that it can extract those rules.}
%including invisible rooms and doors.
%, which we obtained from real-estate listings in a production pipeline.
%Reconstructed floorplans would allow further applications in the real-estate and construction domains (e.g., building code verification and property value assessment) as well as virtual and augmented reality industries (e.g., digital content creation).
%media content creation applications in augmented and virtual reality industries.
%, and more. 
\begin{comment}
\begin{figure}[htb]

    \includegraphics[width=\textwidth]{figures/tea.pdf} 
    \captionof{figure}{The paper presents a new extreme floorplan reconstruction task. An input is a partial floorplan inferred or curated from panorama images of a house/apartment. An output is a complete floorplan. A technical challenge is the hallucination of invisible rooms and doors. %The paper presents a new benchmark and a neural architecture for a solution.}
    }
    \label{fig:my_label}
  \end{figure}  
\end{comment}
  
The technical challenge lies in the restoration or ``hallucination'' of invisible rooms and doors.
%, which requires an understanding of the architectural rules of house layout.
%and hence ambiguity, learn rules of house layout
Inferring invisible image or geometry data has been studied in computer vision in the context of image inpainting~\cite{inpaint2}, illumination inference~\cite{neural_illumination}, amodal segmentation\cite{amodal, amodal2},
and surface reconstruction~\cite{xu2019disn}. To our knowledge, this paper is the first to tackle the hallucination of architectural components, such as rooms and doors, at the scale of an entire house.

%To our knowledge, this paper is the first to tackle the reconstruction of invisible structure at the level of architectural components

Our contribution is three fold: 1) A floorplan restoration task;
2) A benchmark with 701 houses with ground-truth vector floorplan images; and 3) A neural architecture, whose cascading Transformer decoders restore an entire floorplan, including invisible rooms/doors. Qualitative and quantitative evaluations demonstrate the effectiveness of our approach over the existing techniques. We will share our code, and data.

%Concretely, the input of the task is a set of posed panos, where the output is a complete floorplan including even unobserved rooms and doors.
%This extreme floorplan reconstruction exhibits a few fundamental challenges to existing reconstruction techniques. First, many rooms are simply not observed. The approach needs to infer unobserved rooms and also their connections (\ie, doors). Second, only images are given as input and 3D information is very noisy. Requires to exploit architectural regularities more.

%This paper introduces a new problem while expanding an existing extreme SfM dataset for the problem, presents a novel neural architecture, and presents the best results in comparison to existing structured reconstruction problems.

\section{Related Work}

%\yasu{Move inpainting and image restoration into "content hallucination" section. Change floorplan restoration into floorplan reconstruction. Our story is that floorplan reconstruction has been studied a lot in research. In industry, restoration has a big demand as in ricoh,Zillow,magicplan, but has not been tackled. This paper addresses this new task.}

%\subsection{Dataset}
We review related work in 1) floorplan reconstruction,  2) content hallucination, and 3) extreme pose estimation.

% zind , structed3d, rplan, InteriorNet,SUNCG,House3D,SceneNet, Habitat Matterport Dataset, 

\mysubsubsection{Floorplan reconstruction} Floorplan reconstruction has a long history in computer vision with strong ties to the real estate and construction industries. With the success of commodity depth sensors, current methods reconstruct floorplans from 3D point clouds, often via a combination of deep neural networks and optimization~\cite{floornet,monte-floor,pintore2020state, shabani2022housediffusion}. The emergence of panorama cameras meets the growing demand for floorplan reconstruction from images alone~\cite{6909481}. The Zillow indoor dataset provides panorama images and ground-truth floorplans obtained with manual interactions~\cite{zind}. Although floorplan reconstruction has been extensively studied in recent years, there is still significant demand among companies such as Ricoh, Zillow, and MagicPlan~\cite{MagicPlan} for floorplan restoration from real production data collected by real users, despite the incompleteness and sparse coverage of such data. However, most of these companies rely on human annotators to perform the task. This paper presents a novel contribution by addressing the challenge of fully automated floorplan restoration from real production data collected by real users, despite its incompleteness and sparse coverage.

\mysubsubsection{Content hallucination} Techniques such as image inpainting, scene completion, image restoration, and object removal has created realistic structures and textures in missing areas~\cite{removal,inpaint10,yu2019free, MAT,restore, repaint, hays2007scene}.
%Previous research has created realistic structures and textures in missing areas using techniques such as image inpainting, scene completion, image restoration, and object removal~\cite{removal,inpaint10,yu2019free, MAT,restore, repaint, hays2007scene}. In addition,
Neural illumination has been used to infer a spherical illumination image from a single perspective image, while implicit neural surface representation can infer an entire object surface from a single image~\cite{song2019neural, xu2019disn}.
Moreover, Room layout estimation often infers invisible wall layouts behind objects~\cite{AtlantaNet,horizon,dulanet}. This paper infers invisible architectural structures at a house scale (i.e., rooms and doors).

%{\color{blue} Also there are works on indoor reconstruction in presence of occlusion like~\cite{pintore2020state, AtlantaNet} where  the goal is reconstruction 3D models of real-world indoor scenes from captured data, and other works that tackle room layout reconstruction such as ~\cite{horizon, dulanet, led2net} .}
%, reconstructing a complete floorplan.
%geometric structures of invisible rooms and their connections (i.e., doors) and reconstructs a floorplan.

\mysubsubsection{Extreme pose estimation}
Given two RGB-D images with little to no visual overlaps, relative pose estimation is possible by inferring and aligning complete scene structures~\cite{yang2019extreme}. Room-layout estimations are registered to complete a floorplan~\cite{lin2019floorplan}.
Extreme structure from motion (SfM) algorithm estimates the camera poses of panorama images with little to no visual overlaps by learning spatial arrangements of architectural components~\cite{shabani2021extreme, hosseini2023puzzlefusion}.
Instead of camera pose estimation, this paper takes aligned panorama images as a partial floorplan model and reconstructs a complete floorplan, including invisible rooms and doors.

%and reconstructs a complete floorplan, while inferring invisible rooms and doors.

%reconstructs a floorplan given partial reconstruction 
%aligned panorama images

%tackles an ``extreme floorplan reconstruction'', whose problem is to take incomplete floorplan reconstruction (either inferred or manually curated from posed panorama images) and infer a complete floorplan including invisible rooms and doors.

%the panorama images with camera poses and reconstructs a floorplan, while the challenge this time is to infer invisible structures (rooms and doors) by learning regularities of 

%This paper follows E-SfM and reconstructs

%UT austin (extremal pose estimation via hybrid representation) \cite{yang2019extreme}
%Hong Kong (Wenping) eccv20? \cite{lin2019floorplan}
%Amin's iccv2021 \cite{shabani2021extreme} (Look at Amin's related work for more papers)

%Researchers have made progress in the task of ``extreme'' pose estimation~\cite{utaustin,hongkong,amin}, where visual overlaps are too minimal for traditional structure-from-motion~\cite{rome_in_day} and multi-view reconstruction techniques~\cite{towards_internet_scale}. Shabani et al. in particular proposed an extreme structure-from-motion algorithm for sparse indoor panoramas. This paper tackles an extreme floorplan reconstruction problem, which takes aligned panoramas and reconstructs a floorplan.

%\subsection{Structured Geometry Reconstruction}

%\subsection{Layout Generation}

%\subsection{Structure from Motion}

%\subsection{Image Hallucination}

\section{Floorplan Restoration Problem} %Dataset}

%\mysubsubsection{Dataset}
%Do not claim "new" too much. Rather say that we expand the dataset introduced in Amin's... 

%The section explains the input, the output, the dataset, and the metrics.

\begin{figure}[t]

    \centering

    \includegraphics[width=\linewidth]{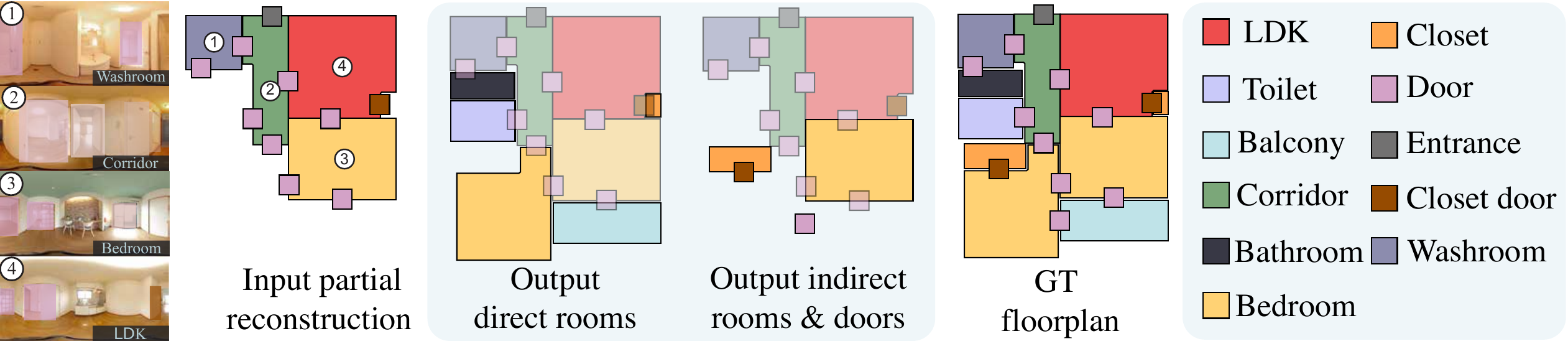}
    \vspace{-0.5cm}
    \caption{An extreme floorplan reconstruction dataset consists of 701 single-floor apartments/houses with ground-truth vector-graphics floorplan images. The technical challenge lies in the reconstruction of invisible rooms and doors.}
   \label{fig:dataset} 
\end{figure}

%\begin{table*}[tbh]

%\end{table*}

%\end{table*}

\mysubsubsection{Input}
The input is a partial floorplan reconstruction, consisting of rooms and doors, each of which is a 14-channel $800\times 800$ segmentation image in a top-down view. 
The raw input data is a set of panorama images in an equirectangular projection with a camera pose, a room layout, door/window detections, and a room type.
Panoramas are acquired with a mono-pod and have roughly the same height from the floor, which allows us to produce an input partial floorplan as a 14-channel image from room layout estimations and camera poses.
There are 10 room types (living room, kitchen, western style room, bathroom, balcony, corridor, Japanese style room, washroom, toilet, and closet) and 4 door types (standard-door, entrance-door, closet-door, and open-portal).
A door is represented as a 2x2 pixel region by the annotators. For the extreme-SfM reconstructions, we identify the door center and replace it by a 2x2 pixel region.
Input data come from an extreme Structure from Motion algorithm~\cite{shabani2021extreme} or human annotators.
%
%Note that a panorama is in an equirectangular projection and Manhattan-rectified (i.e., an image center is aligned with a Manhattan direction).
%Figure~\ref{fig:dataset} shows an example data with 4 panoramas, forming an ``input partial reconstruction'' of 4 rooms with 10 doors by a human annotator.
%is used to product an input partial floorplan as a 14-channel image.
%
%A camera pose is given as a 2D camera position on a horizontal plane (as translation), a constant camera height from the floor, and a 1D heading angle (as rotation), lueecause 1) a panorama is gravity-rectified, which is easy with a use of IMU or vanishing-point detection~\cite{shabani2021extreme}; and 2) Panoramas are acquired with a mono-pod and have roughly the same height from the flat floor.
%, because 1) a panorama image is gravity-rectified; and 2) panoramas are acquired with a mono-pod and have roughly the same height from the flat floor.
%Note that a panorama is in an equirectangular projection and Manhattan-rectified (i.e., an image center is aligned with a Manhattan direction).}
%
%each panorama image is associated with a room layout, door/window detections, and a room type information. 
%as an extra input information.
%(See the next ``dataset'' section for more details).
%
At testing, we fit an axis-aligned bounding box to the room/door masks in a 14-channel image, uniformly scale to fit at the center of a $200\times 200$ square, then add a padding of 300 pixels around, resulting in an $800\times 800$ image.
At training, we use only ground-truth (GT) samples.We uniformly scale an image to fit at the center of a $100\times 100$ square and add a padding of 76 pixels all around to make a $256\times 256$ image.
We apply the same augmentation process in DETR~\cite{detr} (i.e., cropping and resizing), followed by 1) 50\% chance of horizontal flipping and 2) 50\% chance of rotation by 90 or -90 degrees.

\mysubsubsection{Output}
The output is a complete floorplan in a similar format as the input (i.e., a component-wise $800\times 800$ raster mask).
Figure~\ref{fig:dataset} shows a sample, where
the house has 10 rooms and 12 doors. 10 doors are visible and given in the input, leaving 2 ``invisible doors'' to be reconstructed. 4 rooms are visible, leaving 6 ``invisible rooms'' to be reconstructed; 1) 5 of which are adjacent to the input reconstruction and are dubbed ``direct invisible rooms''; and 2) the last of which is dubbed ``indirect invisible room'' (\eg, an invisible closet in an invisible room). 
%
%, that is, an instance-aware segmentation mask for a room or a door. A room needs to be classified to one of the ten room types. Similarly, a door needs to be classified to one of the four door types
%We divided room types to 10 types: living room, kitchen, Western style room, bathroom, balcony, corridor, Japanese style room, washroom, toilet and closet and  4 connection types: room-door-room , room-door-outside, room-door-closet and room-frame-room.
Note that the output of our neural network is a raster floorplan image, which is converted to a vector-graphics floorplan by post-processing.

%Note that our system produces a vector-graphics floorplan. However, vector reconstruction is challenging, and we define the task output to be a raster floorplan for evaluation.

%Ideally, an output floorplan should be in a vector-graphics format, but vector reconstruction is a lot more difficult.
%than raster reconstruction, where
%We define the task output to be a raster floorplan for evaluation, as our task is already challenging.

%which our proposed system produces. However, due to an extreme challenge of the task, we define our output to be a raster floorplan for evaluation.

%due to the challenge, a floorplan is represented as a component-wise raster segmentation mask

%Note that an output should ideally be a vector-graphics floorplan image, which the proposed approach in this paper also produces. However, standard metrics for vector-graphics floorplan reconstruction~\cite{floornet,monte_floor} are too harsh for extremal floorplan reconstruction. 
%this task is extremely challenging and we use a raster floorplan image

%which should ideally be a vector-graphics image but is evaluated as a raster image due to the challenge.

%set of ``invisible'' rooms and doors. In Fig.~\ref{fig:dataset}, five rooms 

\mysubsubsection{Dataset}
%Extreme floorplan reconstruction 
The dataset consists of 701 houses with 
%. The raw data are 
2,355 panorama images captured by Ricoh Theta series in a production pipeline.
The number of panoramas per house ranges from 1 to 7.
Each house has a GT floorplan image, converted to a vector format by manual annotation. Input partial floorplans are inferred by extreme Structure-from-Motion system~\cite{shabani2021extreme} or created from the GT floorplans ( \ie, dropping rooms that do not contain panorama centers).
Randomly sampled 651 houses are used for training while 50 houses are used for testing.
To evaluate the robustness across different datasets, we created a synthetic one from the widely-used RPLAN dataset~\cite{rplan}, while dropping some rooms/doors. We refer to the supplementary for more details on the datasets and comparison with other available datasets.

\mysubsubsection{Metrics} 
We use precision/recall metrics of a floorplan reconstruction paper~\cite{floor_sp} for missing
%The main goal is to restore missing rooms types/shapes and location so we precision/recall metrics of a floorplan reconstruction paper~\cite{floor_sp}.
rooms.~\footnote{We borrow the room metrics but not the corner/edge metrics that are a bit too harsh in our setting.
 %, which require a vector-graphics floorplan and are too harsh in our setting.
 } When the input is a damaged GT, there is no need to align reconstructions. 
We match reconstructed component-wise room masks with the GT, and calculate the precision/recall. We declare that a reconstructed room matches a GT, when the room types match, and the intersection over union (IoU) score is above 0.7, we provided result for other thresholds in the supplementary material. A GT room is matched at most once. In practice, we greedily find matches by: 1) Identifying the match with the highest IoU score; 2) Removing the matched pair; and 3) Repeat. The process is exactly the same for the door metrics except that the IoU threshold is 0.5, because door segments are small.
When the input results from the extreme-SfM system, we need to align the reconstruction with the GT before calculating the metric. We exhaustively try all possible translations (at the granularity of a pixel in a top-down image space), then use the result with the best F1 score.
We also have reported LPIPs~\cite{lpips} and FID~\cite{fid} in ther supplements.

\section{Neural Floorplan Restoration}
%Structure-Hallucinating Transformer Cascades}
\label{sec:method}

\begin{figure*}[tb]
    \centering
    \includegraphics[width=0.90\textwidth]{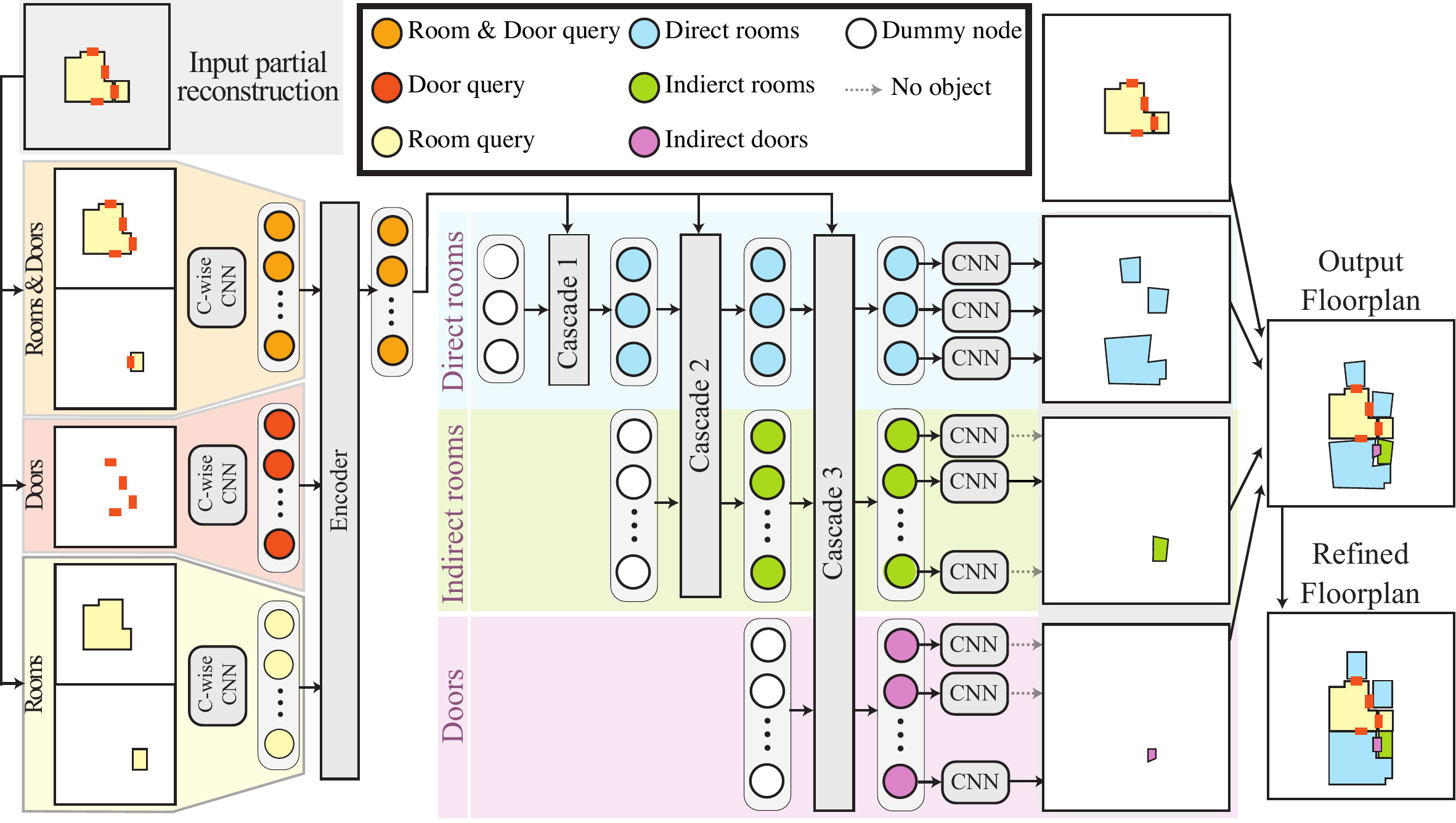}
    \caption{System overview. Category-wise CNNs with a Transformer block encodes an input partial reconstruction into a set of embedding vectors. The cascading Transformer decoders reconstruct invisible rooms and doors in three steps}
    \label{fig:system}
\end{figure*}

Our end-to-end neural architecture consists of a CNN/Transformer based encoder and cascading Transformer decoders. The architecture takes a  partial reconstruction and produces a complete floorplan as component-wise raster segmentation masks, which is refined to a vector-graphics floorplan (See Fig.~\ref{fig:system}). The section explains the key ideas and the design choices. We refer to the supplementary for the full architecture specification.

\subsection{Visible room/door encoder}
Convolutional neural networks (CNNs) with a Transformer encode an input partial reconstruction as a set of $(W\times H\times 14)$ images into a $(\frac{W}{32} \times \frac{H}{32}\times 256)$ feature map in two steps. At testing, $W=H=800$. At training, $W$ and $H$ depend on the augmentation and are around 600.
In the first step, a standard Res-Net~\cite{resnet} (Res-Net-50) processes input images per architectural component category in three branches (i.e., room, door, or both). For example, in the room-branch, we feed each room-image to Res-Net, take the last layer of the last conv-block $(\frac{W}{32}\times \frac{H}{32}\times 2048$), and apply a 1x1 convolution to change the depth to 256. Element-wise maximum across all the rooms results in a $(\frac{W}{32}\times \frac{H}{32}\times 256)$ feature map.
 
We treat the output as $(\frac{W}{32}\times \frac{H}{32})$ tokens with 256-dim embedding for the next Transformer block. ResNet produces the same number of tokens from the other two branches.
The both-branch is the same as the room-branch except that the input image of a room contains masks of its incident doors. In the door-branch, the input is a $(W\times H\times 14)$ image containing all the door masks (instance unaware).
In the second step, a self-attention block from standard Transformer~\cite{transformer} (6 layers w/ 8 heads) takes the tokens from the three branches.

Following the work by Zhou et al.~\cite{informer}, we add standard frequency position encoding to distinguish the branch-type and the X/Y position in the feature map:
\begin{eqnarray}
&&\overrightarrow{f_{xy}} \leftarrow \overrightarrow{f_{xy}} + \left[\overrightarrow{P}_{128}(x), \overrightarrow{P}_{128}(y)\right] + \overrightarrow{P}_{256}(type)\\
&&\overrightarrow{P}_{d}(t) = \left[ \{\cos{(10^{8i/d} t)}\}, \{\sin{(10^{8i/d} t)}\} \right], \ (i=1,2, \cdots d/2).
%\omega_i = 10000^{2i/ d_{\text{model}}}\\
\end{eqnarray}
$\overrightarrow{f_{xy}}$ denotes the 256-dim feature vector of a token at position $(x, y)$, where $x\in\left[1, \frac{W}{32}\right]$ and $y\in\left[1, \frac{H}{32}\right]$.
$\overrightarrow{P}_d(t)$ is a $d$-dimensional standard frequency position encoding.
$(type)$ is a scalar,
%(1000, 1001, or 1002),
indicating the branch-type (both=1001, room=1002, and door=1003).
%, which we represent as an integer (
The tokens from the ``both'' branch are passed to the decoder. 

\subsection{Invisible room/door cascading decoders}
Three cascaded transformer decoders reconstruct direct invisible rooms, indirect invisible rooms, and invisible doors (See Fig.~\ref{fig:dataset}). 

%For decoder we cascade three decoder together to get the results, we call them : Direct invisible room decoder, Invisible room decoder, and Invisible room/door decoder.

\mysubsubsection{Direct invisible room decoder} Direct invisible rooms are behind the doors detected in the panoramas. Let $N_{vis\cdot d}$ be the number of detected doors, which is the expected number of direct invisible rooms. Following DETR~\cite{detr}, we pass $N_{vis\cdot d}$ query tokens with learnable embeddings to a self-attention block, while feeding the tokens from the encoder via cross-attention. 
Each query token will contain an embedding of a direct invisible room to be reconstructed.
The number of detected doors (i.e., query embeddings) varies. We prepare 20 embedding vectors, which is large enough during training.
An output embedding at a query token is used to 1) classify the room-type as a 15-dim vector by a fully connected layer with soft-max; 2) regress the bounding box parameters (i.e., the center, the width, and the height) in the normalized image coordinate ($x,y\in [0,1]$) by a 3-layer MLP with ReLU (hidden dimension 256); and 3) estimate a binary segmentation mask by the panoptic segmentation head in DETR~\cite{detr}.
Note that the type classification labels consist of 10 room types, 4 door types, and ``no room/door". The last label indicates that nothing is reconstructed. At testing, we use the category with the greatest value and keep pixels above the positive value for segmentation.
The bounding box is used for a loss during training, but is not during testing.

\mysubsubsection{Indirect invisible room decoder}
The second cascaded decoder reconstructs indirect invisible rooms via query tokens while passing the encoder tokens via cross-attention in the same way as the first cascade. The only difference is that the self-attention block also incorporates the direct room tokens from the first cascade. 
Both the direct and the indirect room tokens are used to predict the room-type, the bounding box parameters, and the segmentation mask by exactly the same network modules with the same loss functions.
We assume 15 indirect invisible rooms at maximum and pass 15 query tokens with learnable embedding.

%Assuming that there are at most 15 indirect invisible rooms, the second decoder takes the tokens from the first cascade as well as 15 new query tokens again with learnable embeddings. The same shared feed-forward network, a 2-layer MLP, and a 3-layer convnet-decoder produces the bounding box, a room type, and a room binary segmentation mask. Note that the encoder tokens are passed via cross-attention again in this cascade.

%takes each output embedding of a query token, and uses a 2-layer MLP to regress the bounding box coordinate and classify the room-type. A 3-layer 

%Then we use a second decoder to predict direct and indirect rooms,
%The second decoder similar to  first to decoder receives output of encoder and query embedding, but for object queries we have $N_{direct} +15$ where 15 is bigger that number of indirect missing rooms. then instead of randomly initializing our object queries, We initialized the first $N_{direct}$ nodes with output of direct invisible room decoder's outputs embedding. We processed output boxes/labels/masks for all rooms from output of this decoder same as previous decoder, and  matching and Hungarian loss is being applied on this prediction with all rooms boxes/labels and masks.
%Encoder has 8 attention heads, there is 6 encoder layers.

\mysubsubsection{Invisible door decoder}
The third decoder reconstructs invisible doors via query tokens in the same way.
%as the other cascades. 
The difference is that self-attention incorporates all the tokens (direct rooms, indirect rooms, and doors).
A complete floorplan is reconstructed after the cascade.
%The room/door types and the segmentation masks from the third cascade are used to complete the output floorplan.

%Assuming that there are at most 15 invisible doors, the third decoder takes direct-room tokens and indirect-room tokens, while adding 15 new query tokens again with learnable embeddings.

%This decoder, not only predicts rooms but also predict doors.
%The third decoder also receives encoder output and query embedding, in this decoder number of queries are equal to $N_{direct} +15+ 15$, then we initialize first $N_{direct} +15$ input  queries with previous decoder output, output of this decoder is being processed same a  last two,  and  matching and Hungarian loss is being applied on this prediction with all rooms and doors boxes/labels and masks.
%
%For mask head, as in our problem room and door may have overlap instead of pan optic segmentation we do instance segmentation number of heads are 8 and number of layers are 6.

%{\color{green}I will do one ablation on that number 15 in supplementary material, doing experiment on:10, 15, 20, 25, 30, but for it is 15 and 15 is the maximum number of missing indirect room, doors.  }

%\mysubsubsection{Mask Head}multi head attention layer  were receives queries from decoder output and key and value from encoder. Then same as DETR we have a feed forward network,and then we reshape output masks to desired shape. 

%For third decoder we have also tried only door decoder, where input is dummy nodes and encoder output and is being trained to only predict doors, the performance was not good.

\subsection{Loss Functions} \label{sec:training}
Our neutral network reconstructs a floorplan in three cascades. In each cascade, we take the reconstructed components, match them with the corresponding ground-truth components, and inject loss functions.
The matching is done in exactly the same way as DETR~\cite{detr}, based on the type classification and the bounding box estimation.
The first cascade is for invisible direct rooms, and we match against only the GT invisible direct rooms. For the second cascade, we match against the GT invisible direct/indirect rooms. For the third cascade, we match against the GT invisible direct/indirect rooms and invisible doors. We descried the loss fuctions in more details in the supplements.
%
%For the first cascade, we match against only the invisible direct rooms.
% (first: invisible direct rooms, second: invisible rooms, and third: invisible rooms and doors)
%
%The rest of the section explains the loss functions and the data pre-processing steps.
%We use the second cascade as an example while exactly the same applies to the other two. The section explains the loss functions and data augmentation steps.
%how to match reconstructed instances with the ground-truth, our loss functions, and data augmentation steps.

%\mysubsubsection{Loss functions}

\subsection{Floorplan refinement and vectorization}
\label{section:refinement}
Cascading decoders reconstruct a floorplan as raster images, where room boundaries are curvy, and door shapes are irregular. We use a floorplan generative model (House-GAN++\cite{housegan} by Nauata \etal) to refine our output and convert to a vector format.
House-GAN++ is designed to produce floorplans from noise vectors, but can also be used to refine a design without changing the overall arrangement by specifying an entire floorplan as an input constraint. House-GAN++ learns a bias to prefer straight shape boundaries and we use the same post-processing in the original work to produce a vector graphics floorplan. Note that House-GAN++ alone, without our constraint fails to infer an accurate floorplan, as shown in the experimental results next. We refer to the supplementary for more details.

\section{Experimental Results}

We have implemented the proposed system using PyTorch 1.10.0 and Python 3.9.7, and used a workstation with a 3.70GHz Intel i9-10900X CPU (20 cores) and an NVIDIA RTX A6000 GPU. To accommodate the varying number of visible doors/rooms, we use a batch size of 1 during training, while accumulating gradients over 16 samples for a parameter update.
%, inorder to have better generalization we apply back propagation per 16 samples
%We train with batch size 1, on a work station We with
We use AdamW optimizer, while setting the learning rate to $10^{-5}$ for the CNN module in the encoder and $10^{-4}$ for the rest of the network (i.e., a Transformer block in the encoder and the three cascading Transformer decoders). We divide the learning rate by 10 after every 100 epochs. We train for 240 epochs, which takes roughly 22 hours with the workstation.
%
%,after training for 240 epochs (22 hours on A6000) on  our network converges, we implemented our method using pytorch 1.10.0 and python 3.9.7.

%We train with AdamW setting initial transformer learning rate to $10^{-4}$  and backbone to $10^{-5}$ and decreasing it by 0.1 every 100 epochs.

\mysubsubsection{Competing methods}
We compare 
%the proposed approach
against the four competing methods. 

\noindent $\bullet$
The first one is one of the state of the art inpainting methods ``MAT'' with a official of implementation~\cite{MAT}.

\noindent $\bullet$
The second one is Mask-RCNN with an official implementation from Meta AI Research~\cite{detectron2}.
%We evaluate two backbones, ResNet-50 and ResNet-101.

\noindent $\bullet$
The third one is Housegan++~\cite{housegan}, requiring a full bubble diagram where rooms are nodes in that bubble diagram to produce the house layout.
However, Housegan++ is not able to predict missing rooms in our problem. We use our model which works the best in missing room prediction to predict missing rooms and create input bubble diagrams for Housegan++. 
In all iterations we pass visible rooms segmentation masks to Housegan++ and from second iteration we also pass 50\% of invisible room predicted segmentation mask from previous iteration to the network too. We continue iterations until $10^{th}$ iteration.

\noindent $\bullet$
The fourth one is DETR via an official implementation~\cite{detr}.
%by the authors~\cite{detr}.
%, again with the same two backbones (ResNet-50 and ResNet-101).

\vspace{0.1cm}
\noindent
These methods take a single image as an input, while our input is a set of 14-channel images (instance-aware). We perform pixel-wise max-pooling over the images and produce a single 14-channel image as their input. Our output is a set of architectural components, each of which is a binary segmentation mask (i.e., a probability distribution over the room/door types) and the bounding box parameters, which Mask-RCNN and DETR directly produce. 

For MAT, the output is a single instance-unaware 14-channel image. We use flood fill algorithm to find connected components,
%using connected component and flood fill algorithm, 
and discard small components whose areas are less than 4 pixels to produce a floorplan. For Mask-RCNN, we set a threshold of 0.6 on the confidence prediction as that was giving the best performance. We vary the parameter and pick the one with the highest average F1 score.
DETR does not require a threshold: A room/door is not generated when the probability of type ``no room/door'' is the greatest. 

%it predicts a type probability including a category "not a room/door". 

%we removed any produced component less than $2\times2$. then we applied opencv connected component to clean up the output results, but it still failed.

%The post-processing requires one hyper-parameter on the detection confidence 
%%%%%%%%%%%%%%%%%%%%%%%%%%%%%%%%%%%%%%%%%%%%%%%%%%%%%%%%%%%%%

For fair evaluation, we have used 1) the same category based weighing in Sect.~\ref{sec:training}
% (Eq.~\ref{eq:weight}) 
for the type classification loss for Mask-RCNN and DETR; 2) the same parameter update schedule (once every 16 samples); 3) the same data augmentation steps; and 4) the same learning rate.%~\footnote{We have two learning rate schedules, one for Transformer and the other for CNN. Exactly the same schedule is used for DETR. Our schedule for the CNN module is used for Unet and Mask-RCNN.}

%DETR uses the same learning rate as ours. UNet and Mask-RCNN training schedules are set to ours for the CNN module.

% used same learning rates  for detr and ours and same learning rate as our CNN for mask-rcnn, 
 
% in all methods we updated our optimizer(back propagate)  every 16 samples, detr and our followed same augmentation on all networks. 
 
% For unet, output as it is not instance segmentation, we used connected components. We used AdamW optimizer in all networks. 
%all seed has been fixed in all methods.

\begin{table}[!tbh]
    \centering
    \caption{The main quantitative evaluations. We compare against MAT, Mask-RCNN ($\mbox{M}_{\mbox{rcnn}}$), DETR, and Housegan++ (H-PP)
    Input partial floorplans are either ground-truth (left) or outputs from the extreme SfM system~\cite{shabani2021extreme} (right), * means refinement    . 
%The precision, the recall, and the F1 score are reported for the rooms and the doors   
%    The left shows cases when the input partial floorplan is GT (manual curation), while the right shows cases when the extreme SfM system~\cite{shabani2021extreme} produces the input partial reconstructions. T
    %
%    Comparison among Mask-R-CNN, DETR and our method with a ResNet-50 and ResNet-101 backbones on our dataset. On left there are results when masks and camera poses and rotations of partial input layout has been given, on right there are results when masks and camera poses and rotations of partial input layout has been derived using sfm.
    %Our method outperform both Mask-R-CNN and DETR. Our method with backbone ResNet-101 shows weaker performance and we believe it is because of over fitting and data shortage 
    }
\scalebox{0.9}{ \begin{tabular}{l|ccc|ccc|ccc|ccc} % \hline\noalign{\smallskip}
  & \multicolumn{6}{c|}{Manual (GT)} & \multicolumn{6}{c}{Extreme-SfM output} \\
  Method & \multicolumn{3}{c|}{Rooms} & \multicolumn{3}{c|}{Doors}& \multicolumn{3}{c|}{Rooms} & \multicolumn{3}{c}{Doors} \\ % \cline{2-13}
       & Pre. & Rec. & F1 & Pre. & Rec. & F1 &Pre. & Rec. & F1 & Pre. & Rec. & F1  \\ \hline
     %FloorSP &      &      &      &       &      &      &      &      &\\
     MAT     &  20.3     &43.2  & 27.6   &   10.6  & 11.3& 11.0 &  18.2 & 36.5 & 24.2 &10.7 &9.1 &9.8\\
     $\mbox{M}_{\mbox{rcnn}}$ &  \color{magenta} 38.1     &24.2      &   29.6   & 12.0&13.4 & 12.7& \color{orange}37.5 & 23.2 &28.6 & 12.3 & 11.4 &11.8\\
     %$\mbox{M}_{\mbox{rcnn}}$(101) & \color{magenta} 39.2     &18.8      &   25.4    & \color{magenta}14.6& 13.9 & 14.3& \color{cyan}39.1 & 22.6 & 28.6 &\color{orange}13.9 &12.6 &\color{magenta}13.2\\
    H-PP  & 33.7&33.1 & 33.4 &10.3 &9.9 &10.1 & 25.0 & 24.4& 24.7& \color{magenta}12.6 &12.2&\color{magenta}12.4 \\
     DETR       & 30.6     &\color{magenta}47.3      & \color{magenta}  37.1   & \color{magenta}13.0& \color{magenta}13.5 & \color{magenta}13.2&   27.1 &\color{orange}41.5 & \color{magenta}32.8  & 11.0& \color{magenta}13.9 &12.3 \\
     %DETR(101)       &  30.7     & \color{orange}50.4      &    \color{magenta}38.1   & 14.2&  \color{orange}16.1 &\color{magenta} 15.1& 26.8 & \color{cyan}46.3 & \color{magenta}33.9 &\color{magenta}13.3 &\color{orange}15.3 & \color{orange}14.2\\
      Ours   & \color{orange}49.2 & \color{orange}50.8 & \color{orange}50.0 & \color{orange}20.2 & \color{orange}19.3 & \color{orange}19.7&\color{magenta}37.1  &\color{magenta}41.0 &  \color{orange}38.9 &\color{orange} 14.0 & \color{orange}18.4 & \color{orange}15.9 \\
%Ours(101)     & \color{orange}46.2 & 41.4 & \color{orange} 43.7 &  \color{orange}15.7 & \color{magenta}14.8 &  \color{orange}15.2 &31.9 & 39.9 & \color{orange}35.4 &10.3 &13.1 &11.5 \\      
        \hline\hline
      
         Ours*  & \color{cyan}{56.2} &\color{cyan}53.1 & \color{cyan}54.6 & \color{cyan}21.0 &\color{cyan}20.4 & \color{cyan}21.1 & \color{cyan}40.6 &  \color{cyan}44.9 &  \color{cyan}42.6 & \color{cyan}15.6& \color{cyan}19.1 & \color{cyan}17.1
    %\noalign{\smallskip}
    \end{tabular}}
    \label{tab:table3}
\end{table}

\begin{table}[h]
\centering
    \caption{Quantitative evaluations on larger  synthetic dataset, RPLAN~\cite{rplan}. We compare against MAT, Mask-RCNN ($\mbox{M}_{\mbox{rcnn}}$), House-GAN++, and DETR. 
    Input partial floorplans are ground-truth. 
%    The left shows cases when the input partial floorplan is GT (manual curation), while the right shows cases when the extreme SfM system~\cite{shabani2021extreme} produces the input partial reconstructions. T
The precision, the recall, and the F1 score are reported for the rooms and the doors.
    %
%    Comparison among Mask-R-CNN, DETR and our method with a ResNet-50 and ResNet-101 backbones on our dataset. On left there are results when masks and camera poses and rotations of partial input layout has been given, on right there are results when masks and camera poses and rotations of partial input layout has been derived using sfm.
    %Our method outperform both Mask-R-CNN and DETR. Our method with backbone ResNet-101 shows weaker performance and we believe it is because of over fitting and data shortage 
    }
 \begin{tabular}{l|ccc|ccc} % \hline\noalign{\smallskip}
  
  Method &  \multicolumn{3}{c|}{Rooms} & \multicolumn{3}{c}{Doors} \\ % \cline{2-13}
     & Pre. & Rec. & F1 & Pre. & Rec. & F1  \\ \hline
     %FloorSP &      &      &      &       &      &      &      &      &\\
     MAT    & 43.5 &\color{orange}52.4      & 47.5 &10.1 & 13.5 &11.5 \\
     $\mbox{M}_{\mbox{rcnn}}$ &  46.4    &37.6     &  41.5  &11.5& 10.3 & 10.8\\
      House-GAN++& 47.6 &45.3& 46.4&\color{magenta}15.8 &12.7&14.0 \\
      DETR     &\color{magenta}  50.5    &\color{magenta}52.1      &  \color{magenta} 51.3   & 15.4& \color{magenta}16.3 & \color{magenta}15.8  \\ Ours    & \color{orange}64.7 & 51.3 & \color{orange}57.2 & \color{orange}20.8&\color{orange} 18.7  & \color{orange}19.1\\  \hline\hline
      Our* &\color{cyan} 65.8& \color{cyan}53.0& \color{cyan}58.7 &\color{cyan}22.9 & \color{cyan}22.3 &\color{cyan}22.6
  
%     \noalign{\smallskip}
    \end{tabular}
    \label{tab:table10}
\end{table}

\begin{figure}[!tb]
    \centering
  \noindent\includegraphics[width=1\textwidth]{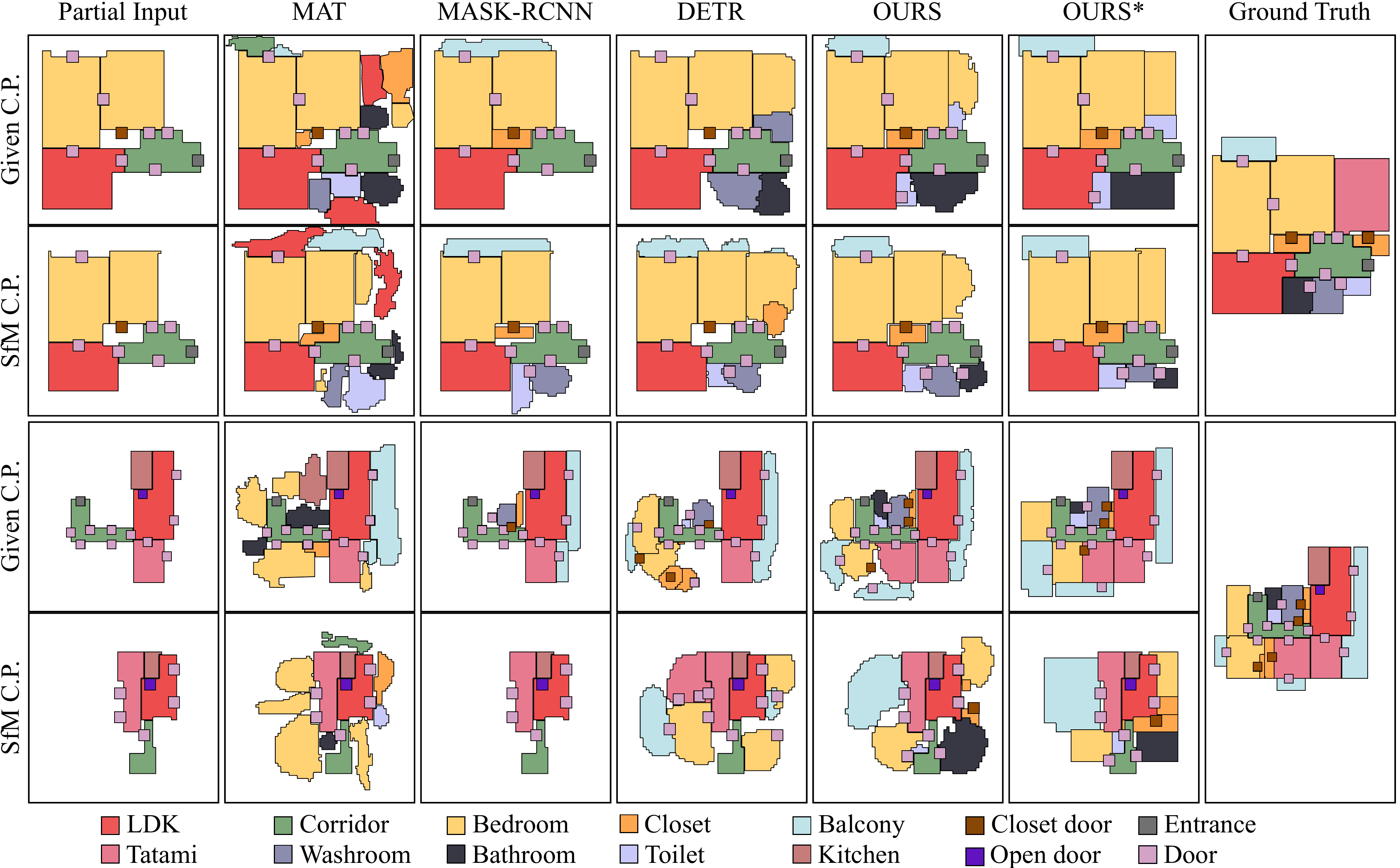} 
    \vspace{-.5cm}
    \caption{Main qualitative evaluations. The figure shows reconstruction results before and after the final refinement for four houses, (Ours, Ours*). For each method, we show both when the input partial reconstruction is from the ground-truth or the extreme SfM system~\cite{shabani2021extreme}.
    At the right, we show both our ground-truth vector-graphics floorplans as well as the original raster floorplan images by an architect.
    %    results, one left there are available panoramas per house, per house top row is when we use gt camera poses and masks for visible rooms, bottom is when we obtain masks and camera poses using SfM. At right end we have ground Truth. 
    }
    \label{fig:fig4}
\end{figure}

\begin{table*}[!tb]
    \centering
    \caption{ Ablation study on the category-wise CNN encoder. The left three columns shows how the input partial reconstructions are passed to each of the three CNN branches. (m) indicates that a branch receives multiple images (instance-aware) as an input. (s) indicates that a branch receives a single image (instance-unaware) that contains masks of all the instances. No-mark indicates that a branch is not used.
    %
    %The table shows not only C-wise cnn imporve the performance, having separate cnn per room and per room-door types can increase the performance as network  while as doors are small instances, having them all together pass results in better door detection.
    }
    \scalebox{0.88}{\begin{tabular}{ccc|ccc|ccc|ccc|ccc} %  \hline\noalign{\smallskip}
      \multirow{3}{*}{\rotatebox[origin=c]{90}{both}} &  \multirow{3}{*}{\rotatebox[origin=c]{90}{room}} &  \multirow{3}{*}{\rotatebox[origin=c]{90}{door}} &  \multicolumn{3}{c|}{Direct Rooms} & \multicolumn{3}{c|}{Rooms}  & \multicolumn{3}{c|}{Rooms} & \multicolumn{3}{c}{Doors}  \\ %
      
    &  &  & \multicolumn{3}{c|}{(First Cas.)} & \multicolumn{3}{c|}{ (Second Cas.)}  & \multicolumn{3}{c|}{(Full Pipeline)} & \multicolumn{3}{c}{(Full Pipeline)} \\ 
      & & & Pre.& Rec.& F1 & Pre.& Rec. & F1 & Pre.& Rec. & F1 & Pre. & Rec. & F1 \\ \hline 
%    \multicolumn{3}{c|}{Cascades}  & \multicolumn{3}{c|}{Mask } & \multicolumn{3}{c}{Bounding Box}\\ 
    s &   &       &  56.3 &    62.6    &    59.3&  40.8&\color{magenta}45.1&  42.8 &\color{magenta}43.8&  46.5 &42.8 & \color{magenta}14.8 & 14.5 &14.6\\
    
    m &   & &  \color{orange}67.8   &  \color{magenta}63.5     &    \color{magenta}65.5  & 34.6&43.9& 38.7 & 35.8  & 48.2  & 41.8 & 6.2 & 7.5 &6.7 \\

    m & m  &     & 63.7    &  62.5     & 63.1  & 35.3&44.6& 39.4 & 36.0   &   \color{magenta}48.6  & 41.3 &  7.1 &  6.9 &  7.0  \\    
        m &   &  s &  60.7  &     61.1   &    60.8  & \color{orange}47.6&\color{orange}45.5& \color{orange}46.3 & 40.6 & 47.6 & \color{magenta}43.9 & 14.1 &\color{magenta}15.6 &\color{magenta}14.8 \\ 
    m & m& m &\color{magenta} 66.2  &\color{cyan}64.7 &\color{orange}65.8&\color{magenta} 46.7 & 44.9 &\color{magenta}45.8 &\color{orange}48.4  &\color{orange}49.2 &\color{orange}48.8& 10.6&11.2& 10.9\\
    
    s & s & s &    57.2   &    54.2    &      55.6 & 37.6&39.7& 38.6 &  38.4 &  40.1  & 39.2 & \color{orange}17.8  &   \color{orange}18.9 & \color{orange}18.3 \\ 
\hline
    m & m & s   & \color{cyan}68.2  & \color{orange}64.2  &\color{cyan}66.1 &  \color{cyan}49.6& \color{cyan}46.6&  \color{cyan}48.0 & \color{cyan}{49.2} & \color{cyan}50.8 & \color{cyan}50.0  & \color{cyan} 20.2   &   \color{cyan}19.3 & \color{cyan}19.7 % \\  
%        \hline\noalign{\smallskip}
    \end{tabular}}
    \label{tab:table4}
\end{table*}

\begin{table}[!tbh]
    \centering
    \caption{Ablation study on the cascading decoders. We turn on and off each of the three cascades (indicated by the left columns) and retrain the entire model.
    The table also reports the metrics of reconstructions by the first and the second cascades. Note that the first (resp. second) cascade reconstructs only direct rooms (resp. direct and indirect rooms), respectively, whose metrics are reported.
    %and the second cascade 
    %Each cascade outputs a floorplan, and the table reports the metrics for such intermediate floorplan reconstructions.
    %
    %Each cascade outputs a floorplan, where the table shows the metrics for the results from each of the three cascades.
    %
    %,  our experiments shows having separate decoder for rooms not only improves room reconstruction performance, but also it importers door prediction. On the other hand having door prediction in general also helps room performance.  
    }
   
    \scalebox{0.88}{\begin{tabular}{ccc|ccc|ccc|ccc|ccc} %  \hline\noalign{\smallskip}
     \multicolumn{3}{c|}{Cascades} & \multicolumn{3}{c|}{Direct Rooms} & \multicolumn{3}{c|}{Rooms}  & \multicolumn{3}{c|}{Rooms} & \multicolumn{3}{c}{Doors}  \\ %
\multirow{2}{*}{\rotatebox[origin=c]{90}{1st}} & \multirow{2}{*}{\rotatebox[origin=c]{90}{2nd}} & \multirow{2}{*}{\rotatebox[origin=c]{90}{3rd}} & \multicolumn{3}{c|}{(First Cas.)} & \multicolumn{3}{c|}{ (Second Cas.)}  & \multicolumn{3}{c|}{(Full Pipeline)} & \multicolumn{3}{c}{(Full Pipeline)} \\ 
       & & &Pre.& Rec.& F1 & Pre.& Rec.& F1 & Pre.& Rec. & F1 & Pre. & Rec. & F1 \\ \hline 
%    \multicolumn{3}{c|}{Cascades}  & \multicolumn{3}{c|}{Mask } & \multicolumn{3}{c}{Bounding Box}\\
       &    &\checkmark & x &  x & x & x & x & x & 36.3 & 41.1   & 38.5 &14.5 & 14.6 &14.5 \\
       & \checkmark & \checkmark  &   x & x & x &39.4 &\color{orange}46.3 &\color{orange}42.6& \color{orange}43.1   & \color{orange}47.0 &\color{orange}44.9 & 14.9&15.2 & 15.0 \\
    \checkmark &    & \checkmark &   \color{orange}67.5&\color{orange} 60.3 & \color{orange}63.6 & x & x & x &36.2   & 41.4& 38.6 &\color{orange}16.3 &\color{orange}16.1 &\color{orange}16.2\\ % \hline
%     \checkmark & \checkmark  & \checkmark  & 68.2   &  64.2   & 66.1 &49.6 & 46.6&48.2 &  49.2    &   50.8 &  50.0  & 20.2& 19.3    &  19.7 \\ \hline \hline 
    \checkmark & \checkmark  &    &    44.8   & 46.3 & 45.5 &\color{orange}40.9&42.0 & 41.5  & 40.9&42.0 & 41.5 & x & x & x \\ \hline
    \checkmark & \checkmark  & \checkmark  & \color{cyan}68.2   &   \color{cyan}64.2   &  \color{cyan}66.1 & \color{cyan}49.6 &  \color{cyan}46.6& \color{cyan}48.2 &  \color{cyan} 49.2    &   \color{cyan} 50.8 &   \color{cyan}50.0  &  \color{cyan}20.2& \color{cyan}19.3    &   \color{cyan}19.7  %  \\          \hline\noalign{\smallskip}
    \end{tabular}}
    \label{tab:table5}
\end{table}

%clip, trim=0.5cm 11cm 0.5cm 11cm  there is a clip option , need to work on correct numbers

\begin{comment}

\begin{figure*}[!p]
    \centering
    \noindent\includegraphics[width=\linewidth]{figures/house.pdf}
    %\vspace{5cm}
    \caption{Reconstructed floorplans before and after the refinement by House-GAN++ and several heuristics. The same refinement process is used for all the methods. The input partial reconstructions are ground-truth in this figure.
    %
    %Refined results for different  competing methods, one left there are available panoramas per house, per house top row is when we before refinement and bottom after refinement using housegan++ . At right end we have ground Truth. 
    \yasu{Cut contents half}
    }
    \label{fig:fig5}
\end{figure*}

\end{comment}

\mysubsubsection{Quantitative evaluations} Table~\ref{tab:table3} provides our main quantitative evaluations, comparing against three competing methods.
The best, the second best, and the third best results are shown in {\color{cyan}cyan}, {\color{orange}orange}, and {\color{magenta}magenta}, respectively. 
Note that the last row is our result with the post-processing refinement step (Sect.~\ref{section:refinement}), which is not an end-to-end system and given as a reference.
%Also we add after Housegan++ result applied to our method to show how Housegan++ can improve performance quantitatively, colors hasn't been for that to emphasize on performance improvements of our method even without Housegan++ comparing to other methods.   
%
Our system achieves the highest F1 score in all the settings. 
Table~\ref{tab:table10} presents results on larger synthetic dataset~\cite{rplan}, where the proposed method outperforms all the other baselines. As it can be seen, shape and type information alone is not sufficient for inpainting methods to handle structured geometry.
Scene graph based generative models such as Housegan++ alone need significant modifications and extra infromation such as full input graph to be adapted to this problem.  Our Transformer cascades provide a more effective solution, performing higher-order reasoning and accurately composing layouts.

%Then we compare our methods on bigger dataset~\cite{rplan} in ~\ref{tab:table10} and our network still outperforms other baselines.

%, where the CNN modules are not critical.
%our approach against the 5 competing methods based on precision, recall and f1 scores. We conduct experiment in two different setups, first when camera poses, rotation and masks for partial layout are given, Second when  camera poses, rotation and masks are being derived using SFM method.

%For Our method, DETR  and Mask-RCNN we used two diffrent backbones: A ResNet-50 and ResNet-101.

%Our method achieves highest f1 score in  both setups. Although Mask-RCNN show slightly better precision in SFM case, its recall is too low which means it does not produce many predictions. 

%To obtains Mask-RCNN, result we use a threshold of 0.6 on prediction scores, We calculate f1 score on threshold varying from 0 to 1 varying by 0.1 and that threshold=0.6 gave us the best  average f1 score in both setup. (%I cab put other threshold in supplements if you think it is needed) 
%In UNET we use~\cite{unet-pytorch} case, we removed any produced component less than $2\times2$. then we applied opencv connected component to clean up the output results, but it still failed.

Table~\ref{tab:table4} provides an ablation study on the category-wise CNN encoder. The bottom row is our final system. In the fifth entry,
% shows a configuration where
the door branch receives multiple images (instance aware), which shows comparable room metrics but worse performance for doors. This suggests that doors are already cleanly separated and an instance unaware single image representation is more efficient and effective.
%captures the key information more effectively. 
%
The last two rows suggest that the room information should be passed as the instance-aware
%multi-image 
representation.

%importance of category wise CNN and our modified encoder, as it can be seen having category wise CNN import the f1 score by 5.8\%. Also, we find out while feeding each room and room-door categories per room to separate CNN can improve the result, for doors it is better to have them as a single image, We thing is may have relation with door size and them being too small comparing to rooms.

Table~\ref{tab:table5} provide an ablation study on the cascading decoders. The middle three rows show that dropping any of the cascades downgrades performance. While the third cascade alone (first row) produces reasonable results, the proposed system with the three cascades achieves the best result in every metric. The last two rows show an interesting phenomena where the third cascade improves the performance of the first two cascades via gradient propagation.

%The table shows that more cascades consistently improves performance, where

%detailed ablation study on cascaded decoders and effect of having door loss on room detection.

\mysubsubsection{Qualitative evaluation} Figure~\ref{fig:fig4} provides the main qualitative evaluations against Mask-RCNN and DETR. ``Ours*" shows the final floorplan models after the refinement by House-GAN++, The refinement process successfully turns raw floorplan reconstructions with many artifacts (\eg, curvy room boundaries and gaps between rooms) into clean floorplan models. Reconstructed floorplans by other different methods before and after the refinement are presented in supplements. 

%Figure~\ref{fig:fig5} shows reconstructed floorplans by different methods before and after the refinement. The refinement process successfully turns raw floorplan reconstructions with many artifacts (\eg, curvy room boundaries and gaps between rooms) into clean floorplan models.
The input and the corresponding ground-truth reveal that this is a challenging reconstruction task, unlike any existing problems. The MAT method produced reasonable pixel-level output but it lacks the ability produce instance-aware output. As a consequence, the results may contain divided or missing rooms. Mask-RCNN is often capable of inferring closets or balconies that are direct neighbors of the input partial reconstruction. However, it fails to infer indirect invisible rooms or doors that are far away from the input reconstruction. With the power of the Transformer, DETR reconstructs indirect rooms and doors more than Mask-RCNN. Nonetheless, our approach infers many invisible structures successfully, in particular when manual (ground-truth) partial reconstructions are given.
Please see the supplementary for more reconstruction examples.

%comparisons against the competing methods. All methods outputs has been refined using Housegan++\cite{housegan}, and fig~\ref{fig:fig5} shows results before and after housegan++  for different baselines, more qualitative results are provided in supplementary material.

\mysubsubsection{Concluding marks}
This paper presents a new floorplan restoration task,
%extreme floorplan reconstruction task,
a new benchmark, and a neural architecture as a solution. The task is challenging and our results are not always satisfactory. Major failure modes are 1) missing rooms; 2) inaccurate room shapes, in particular corridors; and 3) inaccessible rooms without any doors.
We hope that this paper starts an avenue of new research toward an ultimate extreme floorplan reconstruction system capable of reconstructing accurate and realistic floorplans for tens of millions of houses with sparse panorama coverage out there. We will share all our code, model, and data, except the panorama images, for privacy concerns.

\mysubsubsection{Acknowledgement} This research is partially supported by NSERC Discovery Grants with Accelerator Supplements and the DND/NSERC Discovery Grant Supplement, NSERC Alliance Grants, and the John R. Evans Leaders Fund (JELF). This research was enabled in part by support provided by BC DRI Group and the Digital Research Alliance of Canada (alliancecan.ca). We are also thankful to RICOH for sharing the datasets.

\bibliography{egbib}

\begin{thebibliography}{38}
\providecommand{\natexlab}[1]{#1}
\providecommand{\url}[1]{\texttt{#1}}
\expandafter\ifx\csname urlstyle\endcsname\relax
  \providecommand{\doi}[1]{doi: #1}\else
  \providecommand{\doi}{doi: \begingroup \urlstyle{rm}\Url}\fi

\bibitem[Mag()]{MagicPlan}
Magicplan.
\newblock \url{hhttps://www.magicplan.app}.

\bibitem[ric()]{ricoh}
Ricoh360.
\newblock \url{https://www.ricoh360.com/tours/}.

\bibitem[Barnes et~al.(2009)Barnes, Shechtman, Finkelstein, and
  Goldman]{removal}
Connelly Barnes, Eli Shechtman, Adam Finkelstein, and Dan~B Goldman.
\newblock {PatchMatch}: A randomized correspondence algorithm for structural
  image editing.
\newblock volume~28, August 2009.

\bibitem[Cabral and Furukawa(2014)]{6909481}
Ricardo Cabral and Yasutaka Furukawa.
\newblock Piecewise planar and compact floorplan reconstruction from images.
\newblock In \emph{2014 IEEE Conference on Computer Vision and Pattern
  Recognition}, pages 628--635, 2014.
\newblock \doi{10.1109/CVPR.2014.546}.

\bibitem[Carion et~al.(2020)Carion, Massa, Synnaeve, Usunier, Kirillov, and
  Zagoruyko]{detr}
Nicolas Carion, Francisco Massa, Gabriel Synnaeve, Nicolas Usunier, Alexander
  Kirillov, and Sergey Zagoruyko.
\newblock End-to-end object detection with transformers.
\newblock In Andrea Vedaldi, Horst Bischof, Thomas Brox, and Jan-Michael Frahm,
  editors, \emph{Computer Vision -- ECCV 2020}, pages 213--229, Cham, 2020.
  Springer International Publishing.
\newblock ISBN 978-3-030-58452-8.

\bibitem[Chen et~al.(2019)Chen, Liu, Wu, and Furukawa]{floor_sp}
Jiacheng Chen, Chen Liu, Jiaye Wu, and Yasutaka Furukawa.
\newblock Floor-sp: Inverse cad for floorplans by sequential room-wise shortest
  path.
\newblock In \emph{The IEEE International Conference on Computer Vision
  (ICCV)}, 2019.

\bibitem[Cruz et~al.(2021)Cruz, Hutchcroft, Li, Khosravan, Boyadzhiev, and
  Kang]{zind}
Steve Cruz, Will Hutchcroft, Yuguang Li, Naji Khosravan, Ivaylo Boyadzhiev, and
  Sing~Bing Kang.
\newblock Zillow indoor dataset: Annotated floor plans with 360deg panoramas
  and 3d room layouts.
\newblock In \emph{Proceedings of the IEEE/CVF Conference on Computer Vision
  and Pattern Recognition (CVPR)}, pages 2133--2143, June 2021.

\bibitem[Dong et~al.(2022)Dong, Cao, and Fu]{inpaint10}
Qiaole Dong, Chenjie Cao, and Yanwei Fu.
\newblock Incremental transformer structure enhanced image inpainting with
  masking positional encoding.
\newblock In \emph{Proceedings of the IEEE/CVF Conference on Computer Vision
  and Pattern Recognition}, 2022.

\bibitem[et~al.(2022)]{repaint}
Andreas~Lugmayr et~al.
\newblock Repaint: Inpainting using denoising diffusion probabilistic models.
\newblock In \emph{CVPR}, 2022.

\bibitem[Hays and Efros(2007)]{hays2007scene}
James Hays and Alexei~A Efros.
\newblock Scene completion using millions of photographs.
\newblock \emph{ACM Transactions on Graphics (ToG)}, 26\penalty0 (3):\penalty0
  4--es, 2007.

\bibitem[He et~al.(2015)He, Zhang, Ren, and Sun]{resnet}
Kaiming He, Xiangyu Zhang, Shaoqing Ren, and Jian Sun.
\newblock Deep residual learning for image recognition.
\newblock \emph{CoRR}, abs/1512.03385, 2015.
\newblock URL \url{http://arxiv.org/abs/1512.03385}.

\bibitem[Heusel et~al.(2017)Heusel, Ramsauer, Unterthiner, Nessler, and
  Hochreiter]{fid}
Martin Heusel, Hubert Ramsauer, Thomas Unterthiner, Bernhard Nessler, and Sepp
  Hochreiter.
\newblock Gans trained by a two time-scale update rule converge to a local nash
  equilibrium.
\newblock NIPS'17. Curran Associates Inc., 2017.
\newblock ISBN 9781510860964.

\bibitem[Hosseini et~al.(2023)Hosseini, Shabani, Irandoust, and
  Furukawa]{hosseini2023puzzlefusion}
Sepidehsadat Hosseini, Mohammad~Amin Shabani, Saghar Irandoust, and Yasutaka
  Furukawa.
\newblock Puzzlefusion: Unleashing the power of diffusion models for spatial
  puzzle solving, 2023.

\bibitem[Li et~al.(2022)Li, Lin, Zhou, Qi, Wang, and Jia]{MAT}
Wenbo Li, Zhe Lin, Kun Zhou, Lu~Qi, Yi~Wang, and Jiaya Jia.
\newblock Mat: Mask-aware transformer for large hole image inpainting.
\newblock In \emph{Proceedings of the IEEE/CVF Conference on Computer Vision
  and Pattern Recognition}, 2022.

\bibitem[Lin et~al.(2019)Lin, Li, and Wang]{lin2019floorplan}
Cheng Lin, Changjian Li, and Wenping Wang.
\newblock Floorplan-jigsaw: Jointly estimating scene layout and aligning
  partial scans.
\newblock In \emph{Proceedings of the IEEE/CVF International Conference on
  Computer Vision}, pages 5674--5683, 2019.

\bibitem[Liu et~al.(2018)Liu, Wu, and Furukawa]{floornet}
Chen Liu, Jiaye Wu, and Yasutaka Furukawa.
\newblock Floornet: A unified framework for floorplan reconstruction from 3d
  scans.
\newblock In \emph{Proceedings of the European Conference on Computer Vision
  (ECCV)}, September 2018.

\bibitem[Nauata et~al.(2021)Nauata, Hosseini, Chang, Chu, Cheng, and
  Furukawa]{housegan}
Nelson Nauata, Sepidehsadat Hosseini, Kai-Hung Chang, Hang Chu, Chin-Yi Cheng,
  and Yasutaka Furukawa.
\newblock House-gan++: Generative adversarial layout refinement network towards
  intelligent computational agent for professional architects.
\newblock In \emph{Proceedings of the IEEE/CVF Conference on Computer Vision
  and Pattern Recognition}, pages 13632--13641, 2021.

\bibitem[Pintore et~al.(2020{\natexlab{a}})Pintore, Agus, and
  Gobbetti]{AtlantaNet}
Giovanni Pintore, Marco Agus, and Enrico Gobbetti.
\newblock {AtlantaNet}: Inferring the {3D} indoor layout from a single 360
  image beyond the {Manhattan} world assumption.
\newblock In \emph{Proc. ECCV}, August 2020{\natexlab{a}}.
\newblock URL
  \url{http://vic.crs4.it/vic/cgi-bin/bib-page.cgi?id='Pintore:2020:AI3'}.

\bibitem[Pintore et~al.(2020{\natexlab{b}})Pintore, Mura, Ganovelli,
  Fuentes-Perez, Pajarola, and Gobbetti]{pintore2020state}
Giovanni Pintore, Claudio Mura, Fabio Ganovelli, Lizeth Fuentes-Perez, Renato
  Pajarola, and Enrico Gobbetti.
\newblock State-of-the-art in automatic 3d reconstruction of structured indoor
  environments.
\newblock In \emph{Computer Graphics Forum}, volume~39, pages 667--699. Wiley
  Online Library, 2020{\natexlab{b}}.

\bibitem[Shabani et~al.(2021)Shabani, Song, Odamaki, Fujiki, and
  Furukawa]{shabani2021extreme}
Mohammad~Amin Shabani, Weilian Song, Makoto Odamaki, Hirochika Fujiki, and
  Yasutaka Furukawa.
\newblock Extreme structure from motion for indoor panoramas without visual
  overlaps.
\newblock In \emph{Proceedings of the IEEE/CVF International Conference on
  Computer Vision}, pages 5703--5711, 2021.

\bibitem[Shabani et~al.(2022)Shabani, Hosseini, and
  Furukawa]{shabani2022housediffusion}
Mohammad~Amin Shabani, Sepidehsadat Hosseini, and Yasutaka Furukawa.
\newblock Housediffusion: Vector floorplan generation via a diffusion model
  with discrete and continuous denoising.
\newblock \emph{arXiv preprint arXiv:2211.13287}, 2022.

\bibitem[Song and Funkhouser(2019{\natexlab{a}})]{song2019neural}
Shuran Song and Thomas Funkhouser.
\newblock Neural illumination: Lighting prediction for indoor environments.
\newblock In \emph{Proceedings of the IEEE/CVF Conference on Computer Vision
  and Pattern Recognition}, pages 6918--6926, 2019{\natexlab{a}}.

\bibitem[Song and Funkhouser(2019{\natexlab{b}})]{neural_illumination}
Shuran Song and Thomas~A. Funkhouser.
\newblock Neural illumination: Lighting prediction for indoor environments.
\newblock \emph{2019 IEEE/CVF Conference on Computer Vision and Pattern
  Recognition (CVPR)}, pages 6911--6919, 2019{\natexlab{b}}.

\bibitem[Stekovic et~al.(2021)Stekovic, Rad, Fraundorfer, and
  Lepetit]{monte-floor}
Sinisa Stekovic, Mahdi Rad, Friedrich Fraundorfer, and Vincent Lepetit.
\newblock Montefloor: Extending mcts for reconstructing accurate large-scale
  floor plans.
\newblock In \emph{Proceedings of the IEEE/CVF International Conference on
  Computer Vision (ICCV)}, pages 16034--16043, October 2021.

\bibitem[Sun et~al.(2019)Sun, Hsiao, Sun, and Chen]{horizon}
Cheng Sun, Chi-Wei Hsiao, Min Sun, and Hwann-Tzong Chen.
\newblock Horizonnet: Learning room layout with 1d representation and pano
  stretch data augmentation.
\newblock In \emph{The IEEE Conference on Computer Vision and Pattern
  Recognition (CVPR)}, June 2019.

\bibitem[Vaswani et~al.(2017)Vaswani, Shazeer, Parmar, Uszkoreit, Jones, Gomez,
  Kaiser, and Polosukhin]{transformer}
Ashish Vaswani, Noam Shazeer, Niki Parmar, Jakob Uszkoreit, Llion Jones,
  Aidan~N. Gomez, Lukasz Kaiser, and Illia Polosukhin.
\newblock Attention is all you need.
\newblock \emph{CoRR}, abs/1706.03762, 2017.
\newblock URL \url{http://arxiv.org/abs/1706.03762}.

\bibitem[Wu et~al.(2019{\natexlab{a}})Wu, Fu, Tang, Wang, Qi, and Liu]{rplan}
Wenming Wu, Xiao-Ming Fu, Rui Tang, Yuhan Wang, Yu-Hao Qi, and Ligang Liu.
\newblock Data-driven interior plan generation for residential buildings.
\newblock \emph{ACM Transactions on Graphics (SIGGRAPH Asia)}, 38\penalty0 (6),
  2019{\natexlab{a}}.

\bibitem[Wu et~al.(2019{\natexlab{b}})Wu, Kirillov, Massa, Lo, and
  Girshick]{detectron2}
Yuxin Wu, Alexander Kirillov, Francisco Massa, Wan-Yen Lo, and Ross Girshick.
\newblock Detectron2.
\newblock \url{https://github.com/facebookresearch/detectron2},
  2019{\natexlab{b}}.

\bibitem[Xu et~al.(2019)Xu, Wang, Ceylan, Mech, and Neumann]{xu2019disn}
Qiangeng Xu, Weiyue Wang, Duygu Ceylan, Radomir Mech, and Ulrich Neumann.
\newblock Disn: Deep implicit surface network for high-quality single-view 3d
  reconstruction.
\newblock \emph{Advances in Neural Information Processing Systems}, 32, 2019.

\bibitem[Yang et~al.(2019{\natexlab{a}})Yang, Wang, Peng, Wonka, Sun, and
  Chu]{dulanet}
Shang-Ta Yang, Fu-En Wang, Chi-Han Peng, Peter Wonka, Min Sun, and Hung-Kuo
  Chu.
\newblock Dula-net: {A} dual-projection network for estimating room layouts
  from a single {RGB} panorama.
\newblock In \emph{{IEEE} Conference on Computer Vision and Pattern
  Recognition, {CVPR} 2019}, pages 3363--3372, 2019{\natexlab{a}}.

\bibitem[Yang et~al.(2019{\natexlab{b}})Yang, Pan, Luo, Zhou, Grauman, and
  Huang]{yang2019extreme}
Zhenpei Yang, Jeffrey~Z Pan, Linjie Luo, Xiaowei Zhou, Kristen Grauman, and
  Qixing Huang.
\newblock Extreme relative pose estimation for rgb-d scans via scene
  completion.
\newblock In \emph{Proceedings of the IEEE/CVF Conference on Computer Vision
  and Pattern Recognition}, pages 4531--4540, 2019{\natexlab{b}}.

\bibitem[Yu et~al.(2019)Yu, Lin, Yang, Shen, Lu, and Huang]{yu2019free}
Jiahui Yu, Zhe Lin, Jimei Yang, Xiaohui Shen, Xin Lu, and Thomas~S Huang.
\newblock Free-form image inpainting with gated convolution.
\newblock In \emph{Proceedings of the IEEE/CVF International Conference on
  Computer Vision}, pages 4471--4480, 2019.

\bibitem[Yu et~al.(2021)Yu, Zhan, Wu, Pan, Cui, Lu, Ma, Xie, and
  Miao]{inpaint2}
Yingchen Yu, Fangneng Zhan, Rongliang Wu, Jianxiong Pan, Kaiwen Cui, Shijian
  Lu, Feiying Ma, Xuansong Xie, and Chunyan Miao.
\newblock Diverse image inpainting with bidirectional and autoregressive
  transformers.
\newblock \emph{Proceedings of the 29th ACM International Conference on
  Multimedia}, 2021.

\bibitem[Zamir et~al.(2022)Zamir, Arora, Khan, Hayat, Khan, and Yang]{restore}
Syed~Waqas Zamir, Aditya Arora, Salman Khan, Munawar Hayat, Fahad~Shahbaz Khan,
  and Ming-Hsuan Yang.
\newblock Restormer: Efficient transformer for high-resolution image
  restoration.
\newblock In \emph{Proceedings of the IEEE/CVF Conference on Computer Vision
  and Pattern Recognition (CVPR)}, pages 5728--5739, June 2022.

\bibitem[Zhan et~al.(2020)Zhan, Pan, Dai, Liu, Lin, and Loy]{amodal2}
Xiaohang Zhan, Xingang Pan, Bo~Dai, Ziwei Liu, Dahua Lin, and Chen~Change Loy.
\newblock Self-supervised scene de-occlusion.
\newblock In \emph{Proceedings of the IEEE conference on computer vision and
  pattern recognition (CVPR)}, June 2020.

\bibitem[Zhang et~al.(2018)Zhang, Hu, Luo, Zuo, and Wang]{lpips}
Haoran Zhang, Zhenzhen Hu, Changzhi Luo, Wangmeng Zuo, and Meng Wang.
\newblock Semantic image inpainting with progressive generative networks.
\newblock In \emph{Proceedings of the 26th ACM International Conference on
  Multimedia}, MM '18, page 1939–1947, New York, NY, USA, 2018. Association
  for Computing Machinery.
\newblock ISBN 9781450356657.
\newblock \doi{10.1145/3240508.3240625}.
\newblock URL \url{https://doi.org/10.1145/3240508.3240625}.

\bibitem[Zhou et~al.(2021)Zhou, Zhang, Peng, Zhang, Li, Xiong, and
  Zhang]{informer}
Haoyi Zhou, Shanghang Zhang, Jieqi Peng, Shuai Zhang, Jianxin Li, Hui Xiong,
  and Wancai Zhang.
\newblock Informer: Beyond efficient transformer for long sequence time-series
  forecasting.
\newblock In \emph{The Thirty-Fifth {AAAI} Conference on Artificial
  Intelligence, {AAAI} 2021, Virtual Conference}, volume~35, pages
  11106--11115. {AAAI} Press, 2021.

\bibitem[Zhu et~al.(2017)Zhu, Tian, Mexatas, and Doll{\'a}r]{amodal}
Yan Zhu, Yuandong Tian, Dimitris Mexatas, and Piotr Doll{\'a}r.
\newblock Semantic amodal segmentation.
\newblock In \emph{Conference on Computer Vision and Pattern Recognition
  ({CVPR})}, 2017.

\end{thebibliography}

\end{document}

\maketitle
\appendix

%\yasu{As we change the sections, please update here. Use exactly the same phrases here as the section titles. In general, too many sections with similar headings, and I feel that we can merge sections.}
The supplementary document provides details of the datasets (\autoref{section:dataset}), 
%background on other related datasets and more details on our dataset in section~\ref{section:dataset}, 
the full specifications of our neural architecture (\autoref{sec:spec}), 
discussion on computational complexity (\autoref{sec:compute}), details of our loss functions (\autoref{sec:loss_sup}),  additional details on refinement process (\autoref{section:refinment}), results under different IOUs (Figure~\ref{fig:IOU}), and more quantitative and qualitative results (Table~\ref{tab:tablefid} and Figures~\ref{fig:sup5} and ~\ref{fig:sup4}).

%The supplementary document provides background on other related datasets and more details on our dataset in section~\ref{section:dataset}, the full specifications of our neural architecture in Section~\ref{sec:spec},
% more information about loss function in section~\ref{sec:loss_sup} and computation complexity section~\ref{sec:compute} , 
% and section~\ref{section:refinment}, results on different IOUs provided in Figure~\ref{fig:IOU} and more experimental results in Table~\ref{tab:tablefid} and in Figures~\ref{fig:sup4},~\ref{fig:sup5} .

\section{Datasets details}
%\yasu{Structure of this section is a bit mysterious. 1.1 is about other datasets. 1.2 is ours but also about another dataset: RPLAN. Why not also talk about RPLAN in 1.1? Or remove subsections and just write a few paragraphs... Can you fix this section? Think about the proper structure of the presentation here.}

\label{section:dataset}

SUN360 is a popular indoor panorama dataset~\cite{xiao2012recognizing}. 
For floorplans, RPLAN provides sixty thousand synthetic samples in a vector-graphics format~\cite{rplan}. LIFULL HOME's dataset provides five million real samples in a raster format~\cite{raster_to_vector}. For both panorama images and floorplans, Structured3D provides 3,500 synthetic houses/apartments~\cite{Structured3D}.
For real samples, Matterport3D provides 90 houses~\cite{chang2017matterport3d}, while their coverage is extremely dense.

ZIND dataset is the closest to ours with 1,500 real houses/apartments, where the panorama coverage is sparse, and human interventions were required for camera pose estimation and floorplan reconstruction~\cite{zind}.
Nonetheless, ensuring a panorama(s) is taken in every room (except stairs and corridors) makes the setup small-scale and highly controlled. Our data comes from uncontrolled crowd-sourcing, where many rooms are not photographed, posing fundamental challenges to existing techniques.

In our proposed dataset, the number of panoramas per house ranges from 1 to 7. Specifically, (31, 129, 222, 199, 118, 2) houses contain (1, 2, 3, 4, 5, 7) panoramas, respectively. The test set contains (9, 19, 15, 7) houses containing (2, 3, 4, 5) panorama images, respectively.
The number of invisible rooms varies between 1 and 17. The number of invisible doors varies between 0 and 11.

As it has been stated in main paper, in order to solve class imbalance problem we use a weighing constant ($w(i)$) that is inversely proportional to the number of samples in the training set, which is (1.135, 3.7, 0.6, 0.63, 0.47, 1.2, 2.2, 0.76, 0.48, 0.18, 0.763, 0.524, 0.33, 0.7, and 0.1) for (living room, kitchen, bedroom, toilet, balcony, corridor, tatami, washroom. bathroom, closet, closet door, open door, door, entrance, and no-room), respectively. Table~\ref{table:statistics} present more detail on dataset statistic. 

\begin{table}[h]
\centering
\caption{Dataset statistics.
We divide the set into five groups based on the number of rooms (2-5, 6-9, 10-13, 14-17, 18+).
In each group, the table reports (Top) the number of samples; (Middle) the ave/std of the number of three types of rooms; and (Bottom) the ave/std of the number of two types of doors.}
\label{table:statistics}
\begin{tabular}{l|l|ccccc} 

%\noalign{\smallskip}
\multicolumn{2}{c|}{\# of Rooms} & 2-5  & 6-9 & 10-13 & 14-17 & 18+ \\
%\noalign{\smallskip}
\hline
%\noalign{\smallskip}
\multicolumn{2}{c|}{\# of Samples} & 60 & 243 & 298 & 84 & 16 \\
\hline
\multirow{3}{*}{\rotatebox[origin=c]{90}{Room}} & Visible & 2.4/0.9 & 3.5/1.2 & 3.7/1.6 & 3.9/1.5 & 3.4/1.0 \\
                       & Invis. direct   & 1.6/0.5 & 3.7/0.9 & 5.2/1.6 &6.4/2.2 & 7.2/2.4 \\
                       & Invis. indire. & 0.7/0.9 & 1.4/1.4 & 3.7/2.2 & 5.0/2.8 & 8.2/3.2 \\ \hline
\multirow{2}{*}{\rotatebox[origin=c]{90}{Door}} & Visible            & 3.0/1.2 & 7.2/1.6 & 8.4/2.5 & 10.6/4.0 & 11.4/2.3 \\
                       & Invisible          & 0.8/1.0 & 1.5/1.7 & 4.4/3.3 & 5.9/3.2 & 9.1/3.5
\end{tabular}

\end{table}
%Here, we first give more detail on our proposed dataset, then provide some details on the RPLAN dataset. 
To
evaluate the robustness across different datasets, we also used RPLAN~\cite{rplan} datsaset which is a public dataset.
We use Housegan++~\cite{housegan} data parser to parse RPLAN, which results in 60K houses; the number of rooms per house varies between 5 to 8. We divide the dataset into 55K for training and 5K for testing. Using our dataset's statistics, we flag the number of rooms per house as invisible. The number of visible rooms per house varies between 2 to 7, the number of invisible rooms varies between 1 to 6, and the number of invisible doors varies between 0 to 5.

%\section{More qualitative results}

%First in Fig~\ref{fig:sup1}, we show SfM camera pose vs GT camera pose of same samples that we show in figure 5 in main paper, where there we only show GT camera pose version before and after housegan++.

\section{Network Architecture} \label{sec:spec}
\mysubsubsectionA{Category Wise CNN} takes an input image of the resolution $800\times 800 \times 14$. 
For the house with X number of visible rooms, We will have X number of C-wise CNN for rooms category were the output will be $X \times 256 \times 25\times 25$, then we apply max function which result output  of that to be $  256 \times 25\times 25$. Same  will be applied to door and room (both) category. For doors only category there is no need for max, as we give all doors at same time to one C-wise CNN block. Then we concatenate them resulting output of all to be $3\times 256\time 25 \time25$, then we reshape them to get input sequence for encoder to be $1875 \times 256$, let's refer it as \emph{src}.

\begin{table*}[tbh]
    \centering
    \caption{Category Wise CNN (C-Wise CNN) block}
    \label{tab:cnn}
    \begin{tabular}{c|c|c}
    
        Layer name/type &  Specification & Output size  \\ \hline\hline
       Conv1  & $ 7\times7$, 64, stride 2   & $64 \times 400 \times 400$ \\ \hline
       
        \multirow{4}{*}{Conv block2} & $3\times 3$, maxpool, stride 2 &    \multirow{4}{*}{$256 \times 200 \times 200$}\\
       &   \multirow{3}{*}{$\left[\begin{array}{l}1 \times 1,64 \\
3 \times 3,64\\  1 \times 1,256\\
 \end{array}\right] \times 3$}   &  \\
          &  & \\
          &  & \\ \hline
          \multirow{3}{*}{Conv block3} & 
          \multirow{3}{*}{$\left[\begin{array}{l}1 \times 1,128 \\
3 \times 3,128\\  1 \times 1,512\\
 \end{array}\right] \times 4$}   & \multirow{3}{*}{$512 \times 100 \times 100$} \\
          &  & \\
          &  & \\ \hline

            \multirow{3}{*}{Conv block4} & 
          \multirow{3}{*}{$\left[\begin{array}{l}1 \times 1,256 \\
3 \times 3,256\\  1 \times 1,1024\\
 \end{array}\right] \times 6$}   & \multirow{3}{*}{$1024 \times 50 \times 50$}\\
          &  & \\
          &  & \\ \hline
          
            \multirow{3}{*}{Conv block5} & 
          \multirow{3}{*}{$\left[\begin{array}{l}1 \times 1,512 \\
3 \times 3,512\\  1 \times 1,2048\\
 \end{array}\right] \times 3$}   & \multirow{3}{*}{$2048 \times 25 \times 25$}\\
          &  & \\
          &  & \\ \hline
          Conv6  & $ 1\times1$, 256  & $256 \times 25 \times 25$ \\ \hline
    \end{tabular}
\end{table*}

\vspace{0.1cm}
\mysubsubsectionA{Encoder} has 6 layers (See Table~\ref{tab:encoder}), it receives \emph{src}, position, and type embedding as input. As positional embedding shape is $625 \times 256$, we concatenate it by itself three times to produce $1875 \times 256$ positional embedding. Type embedding shape is $3 \times 256 $, we concatenate each embedding related to each category by itself 625 times and concatenate all them together to produce $1875 \times 256$ type embedding. Encoder input is summation of \emph{src} and those two embeddings.

Final output of encoder will be $1875 \times 256$. We only use the ones for room-door (both) category as decoder input which will be $625 \times 256$, lets refer this as  \emph{memory (Mem)}.

\begin{table*}[tb]
    \centering
    \caption{One layer of encoder, we have 6 layers}
    \label{tab:encoder}
    \scalebox{0.9}{
\begin{tabular}{c|c|c|c}
    
        Layer name/type &  input & Specification & Output size  \\ \hline\hline
        
        Multi head attention & src & embed\_dim=256, num\_heads=8 &  $1875 \times 256$ (src2)\\
        
        Dropout &   src2 & dropout=0.1 & $1875 \times 256$ (src3)\\
        
        LayerNorm & src3+src &   normalized\_shape=256 & $1875 \times 256$ (src4)\\ \hline

        Linear1, ReLU & src4 & in\_features=256, out\_features=2048 &  $1875 \times 2048$ (src5)\\ 
        
         Dropout &   src5 & dropout=0.1 & $1875 \times 2048$ (src6)\\
        
        Linear2 & src6&  in\_features=2048, out\_features=256  &$1875 \times 256 $(src7)
 \\        
         Dropout &   src7 & dropout=0.1 & $1875 \times 256$ (src8)\\
        
        LayerNorm & src8+src4 &   normalized\_shape=256 & $1875 \times 256$ (src9)\\ \hline

    \end{tabular}}
\end{table*}

\vspace{0.1cm}
\mysubsubsectionA{Decoder} has three cascaded blocks and 6 layers in each block. Table~\ref{tab:decoder} shows one layer in one block. Let us assume that we have Y dangling doors. then number of queries for first decoder will be Y, second will be Y+15 and third will be Y+30. Lets show each cascade input queries as \emph{tgt}.
The first cascade input is Y dummy queries and encoder output (\emph{Mem}). Then first cascade output will be concatenate with 15 dummy queries the result is input queries for second decoder cascade, second cascade also receives \emph{Mem} as input. For last cascade, second cascade output will be concatenate with 15 dummy queries and same as other two cascades, third cascade also receives \emph{Mem} too.

\begin{table*}[h!]
    \centering
    \caption{One layer of decoder, we have 6 layers, Y is number of input queries, in first cascade it is equal to number of dangling doors, in second it is number of dangling doors+15, and in third number of dangling doors+30}
    \label{tab:decoder}
    \scalebox{0.95}{
    \begin{tabular}{c|c|c|c}
        Layer name/type &  input & Specification & Output size  \\ \hline\hline
        Multi head attention & tgt & embed\_dim=256, num\_heads=8 &  $Y \times 256$ (tgt2)\\
        Dropout &   tgt2 & dropout=0.1 & $Y \times 256$ (tgt3)\\
        LayerNorm & tgt3+tgt &   normalized\_shape=256 & $Y \times 256$ (tgt4)\\ \hline
        
        Multi head attention & Mem, tgt & embed\_dim=256, num\_heads=8 &  $Y \times 256$ (tgt5)\\
        Dropout &   tgt6 & dropout=0.1 & $Y \times 256$ (tgt7)\\
        LayerNorm & tgt7+tgt4 &   normalized\_shape=256 & $Y \times 256$ (tgt8)\\ \hline
        Linear1, ReLU & tgt8 & in\_features=256, out\_features=2048 &  $Y \times 2048$ (tgt9)\\ 
         Dropout &   tgt9 & dropout=0.1 & $Y \times 2048$ (tgt10)\\
        Linear2 & tgt10&  in\_features=2048, out\_features=256  &$Y \times 256 $(tgt11)
 \\        
         Dropout &   tgt11 & dropout=0.1 & $Y \times 256$ (tgt12)\\
        LayerNorm & tgt8+tgt12 &   normalized\_shape=256 & $Y \times 256$ (tgt13)\\ \hline
    \end{tabular}}
\end{table*}
\subsection{Computational complexity} 
 The number of network parameters is 50 million ResNet 50. As a reference, the network size of DETR  [3] resp. 40 million. Note that our network size is not very different from other standard Transformer based architecture. The reason for choosing a batch size of  1 is not due to memory constraints, but to accommodate the varying number of visible doors/rooms per instance during implementation. 
 \label{sec:compute}

\subsection{Loss functions}
\label{sec:loss_sup}
Without loss of generality, we use the second cascade as an example to define loss functions, where reconstructed invisible room instances are matched against the corresponding ground-truth.
%e use the second cascade as an example while exactly the same applies to the other two. The section explains the loss functions and data augmentation steps.
%We first define loss functions assuming that the matching was done: Each reconstructed room instance is either given a corresponding ground-truth room instance or declared as no-match.
%Assuming that reconstructed rooms are matched with the ground-truth rooms,
The loss functions consist of the three terms for the room-type classification, the bounding box parameter regression, and the segmentation mask estimation:
\begin{eqnarray}
L &=& L_{type} + L_{bbox} + L_{seg},\\
L_{type} &=& \frac{1}{|\mathcal{I}_{all}|}\sum_{i\in \mathcal{I}_{all}} - w(i) log (p_i), \\
L_{bbox} &=&\frac{1}{|\mathcal{I}_{match}|} \sum_{i \in \mathcal{I}_{match}} 5 \left\|b_{i}-\hat{b}_i\right\|_{1} - 2\ \mbox{IOU}(b_i, \hat{b}_i), \label{eq:weight} \\
L_{seg} &=& \frac{5}{|\mathcal{I}_{match}|}\sum_{i \in \mathcal{I}_{match}}\left[ L_{dice}(m_i, \hat{m_i}) + \frac{1}{|\mathcal{P}|} \sum_{p\in \mathcal{P}}  L_{focal}(m^p_i, \hat{m^p_i}) \right].
\end{eqnarray}  

\begin{comment}
\begin{equation}
L = L_{type} + L_{bbox} + L_{seg}
\end{equation}
\begin{equation}
L_{type} = \frac{1}{|\mathcal{I}_{all}|}\sum_{i\in \mathcal{I}_{all}} - w(i) log (p_i)
\end{equation}
\begin{equation}
L_{bbox} =\frac{1}{|\mathcal{I}_{match}|} \sum_{i \in \mathcal{I}_{match}} 5 \left\|b_{i}-\hat{b}_i\right\|_{1} - 2\ \mbox{IOU}(b_i, \hat{b}_i) \label{eq:weight} 
\end{equation}
\begin{equation}
L_{seg\_match}(i) = L_{dice}(m_i, \hat{m_i}) + \frac{1}{|\mathcal{P}|} \sum_{p\in \mathcal{P}}  L_{focal}(m^p_i, \hat{m^p_i})
\end{equation}
\begin{equation}
L_{seg} = \frac{5}{|\mathcal{I}_{match}|}\sum_{i \in \mathcal{I}_{match}} L_{seg\_match}(i)
\end{equation}    
\end{comment}

$L_{type}$ is a standard softmax loss with per-category weighing. $\mathcal{I}_{all}$ is a set of indexes of all the reconstructed room instances. With abuse of notation, $p_i$ denotes the classification score of the GT room type (i.e., the room type of the matched GT). $w(i)$ denotes a weight associated with the GT room type to compensate for the imbalance in the training data. Concretely, the weight is set inversely proportional to the number of samples in the training set.

$L_{bbox}$ sums the discrepancies of the bounding box estimation over the matched instances. The first term is the L1 norm of the 4-dimensional bounding box parameter vector (the center, the width, and the height in the normalized image coordinate). $b_i$ denotes the estimated parameter vector and $\hat{b_i}$ denotes the corresponding GT vector.
The second term is the intersection over the union score between the reconstructed and the ground-truth bounding boxes.

$L_{seg}$ sums the discrepancy of the segmentation mask and the matched instance. The first term is the standard dice loss between the estimated mask ($m_i$) and the GT ($\hat{m_i}$). The second term is the average focal loss of the per-pixel mask value. $\mathcal{P}$ denotes the set of pixels in the domain. $m^p_i$ (resp. $\hat{m}^p_i$) denotes the estimated (resp. GT) per-pixel mask value.

%\section{Housegan++ Refinement Detail}
\subsection{Final floorplan refinement}
\label{section:refinment}

Here we give a few more details on our refinement process. RPLAN dataset~\cite{rplan} was initially used to train House-GAN++ but is synthetic and may exhibit incompatible database bias. We use LIFULL HOME'S dataset~\cite{LIFUL} instead. Since the goal is the refinement without overall arrangement changes, we use fully-connected graphs for training and testing. The reconstructed floorplan from our cascading decoder is specified as the input constraint. In the first iteration, we give our last cascade output as inputs mask to Housegan++. The predicted mask of the previous iteration will be input for each node from the second iteration, and we continue this to the 10th iteration. By doing this, as we are limiting Housegan++, masks only get refined and curvy edges will change to manhattan shapes. 

%input{sections/diffrent_iou}
\section{Additional Experimental Results}

\subsection{LPIPS and FID Results}
As discussed in the main paper, FID and LPIPS are not as powerful metrics as much as our main metric to show if our method is successful in our unique task however we show them here for the reference. 

\begin{table}[!tbh]
\centering
    \caption{FID and LPIPS results. We compare against MAT, Mask-RCNN ($\mbox{M}_{\mbox{rcnn}}$), House-GAN++, and DETR with ResNet-50 backbone. Input partial floorplans are ground-truth. }
%    The left shows cases when the input partial floorplan is GT (manual curation), while the right shows cases when the extreme SfM system~\cite{shabani2021extreme} produces the input partial reconstructions. T

%    Comparison among Mask-R-CNN, DETR and our method with a ResNet-50 and ResNet-101 backbones on our dataset. On left there are results when masks and camera poses and rotations of partial input layout has been given, on right there are results when masks and camera poses and rotations of partial input layout has been derived using sfm.
    %Our method outperform both Mask-R-CNN and DETR. Our method with backbone ResNet-101 shows weaker performance and we believe it is because of over fitting and data shortage 
    
 \begin{tabular}{l|c|c} % \hline\noalign{\smallskip}
  
  Method & $\downarrow$ FID & $\downarrow$ LPIPS \\ % \cline{2-13}
  \hline
    
     %FloorSP &      &      &      &       &      &      &      &      &\\
     MAT    & 80.9& 0.245 \\
     $\mbox{M}_{\mbox{rcnn}}$ & 76.0   & 0.241 \\
      House-GAN++& 45.6& 0.220 \\
      DETR     & 79.1&  0.243\\ 
      Ours & 76.8 & 0.242 \\
      \hline\hline
      Our* & 42.4 & 0.214
  
%     \noalign{\smallskip}
    \end{tabular}
    \label{tab:tablefid}
\end{table}

\subsection{Additional results on different IOU-thresholds}

\begin{figure}[!tbh]
    \centering
   \includegraphics[width=\linewidth]{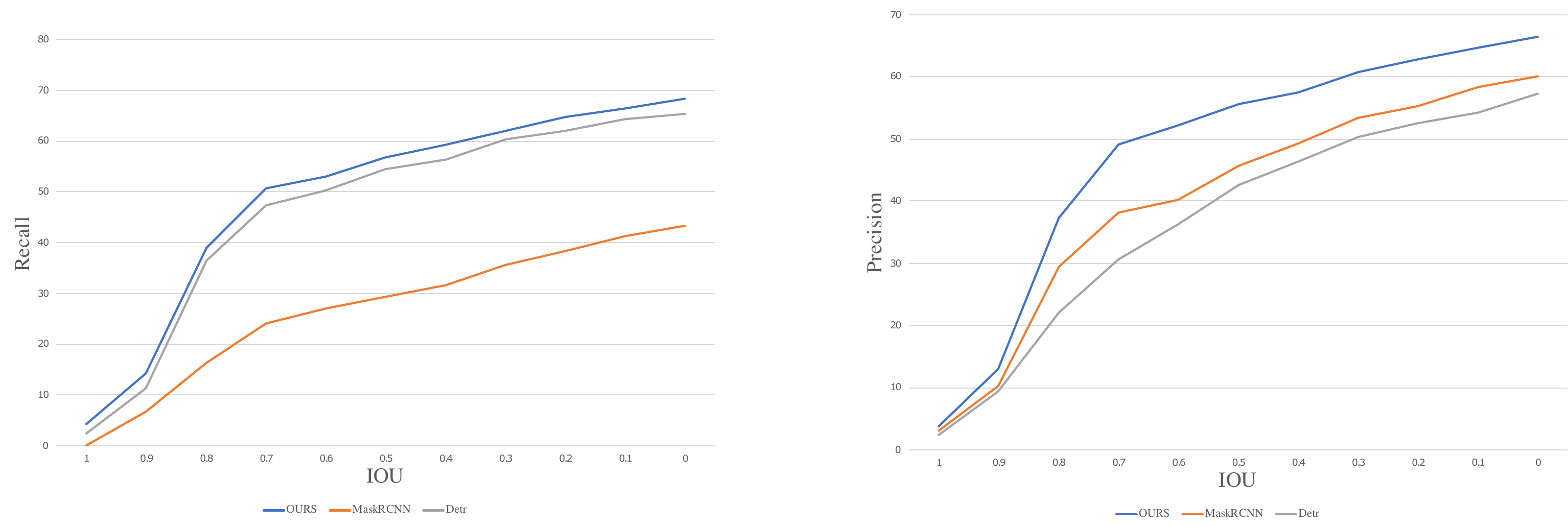}
    \caption{Both the precision and the recall decrease as we increase the IOU threshold in Fig5, which is not what a normal precision/recall curve shows, for example in a detection task. This is because, our plots are based on the IOU threshold on the metric computation, as opposed to a confidence threshold of a detection in a normal precision/recall curve.}
    \label{fig:IOU}
\end{figure}

In Figure~\ref{fig:IOU} we are providing recall and precision graph for different IOUs starting from 0 going up to 1, increasing by 0.1. 
Both the precision and the recall decrease as we increase the IOU threshold. As our plots are based on the IOU threshold on the metric computation, by decreasing the IOU threshold, number of True positive increases in both precision and recall which cause the increase in both of them, in all IOU thresholds our method has the highest performance.

\begin{figure}[!tbh]
    \centering
    \includegraphics[width=0.92\linewidth]{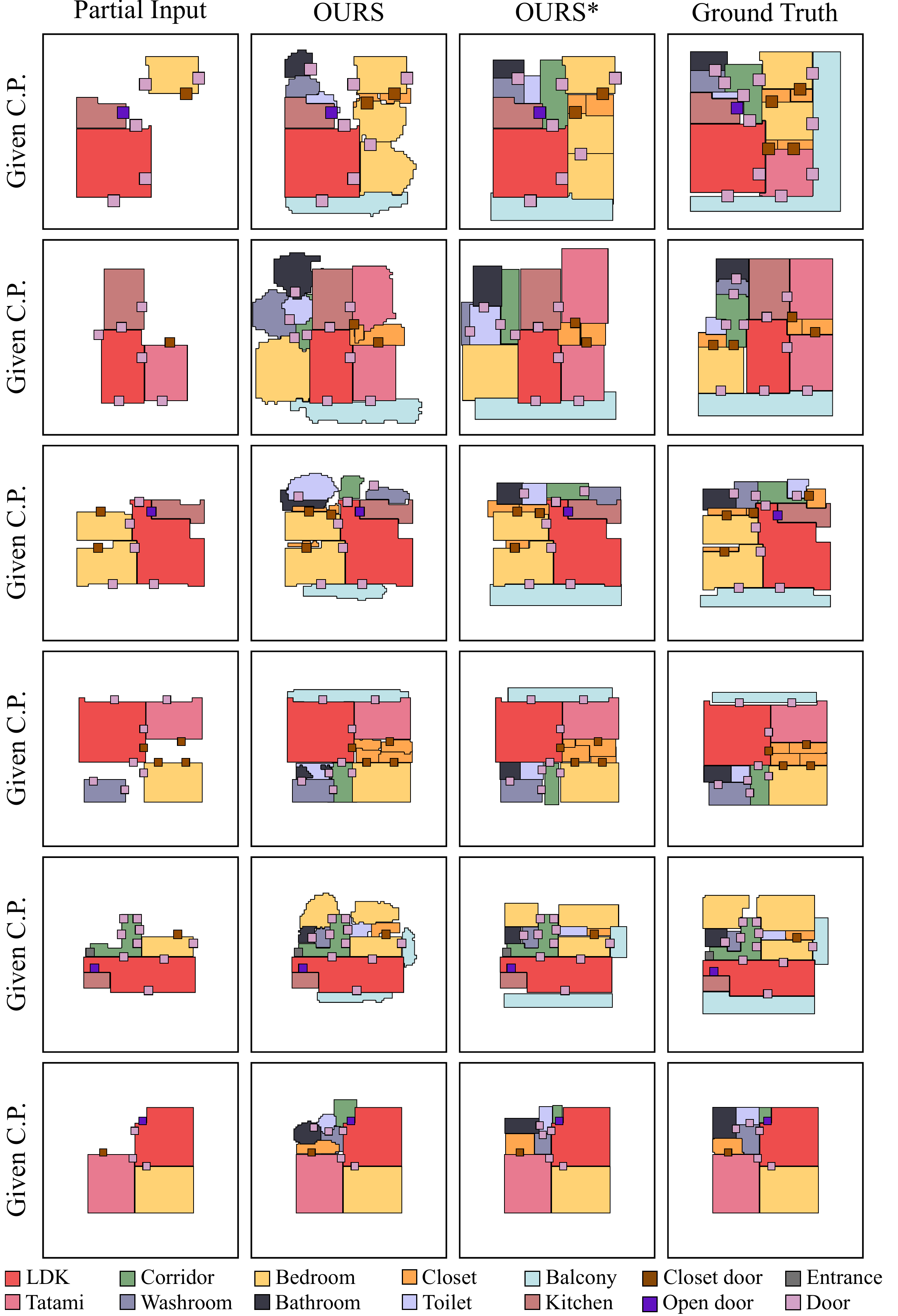}
gr    \caption{More qualitative results. Reconstructed floorplans before and after the refinement for our method by House-GAN++ and several heuristics for more samples. The input partial reconstructions are derived either by using given Camera Pose in this figure.}
    \label{fig:sup5}
\end{figure}
\subsection{More qualitative results}
\begin{figure}[!thb]
    \centering
    \includegraphics[width=\linewidth]{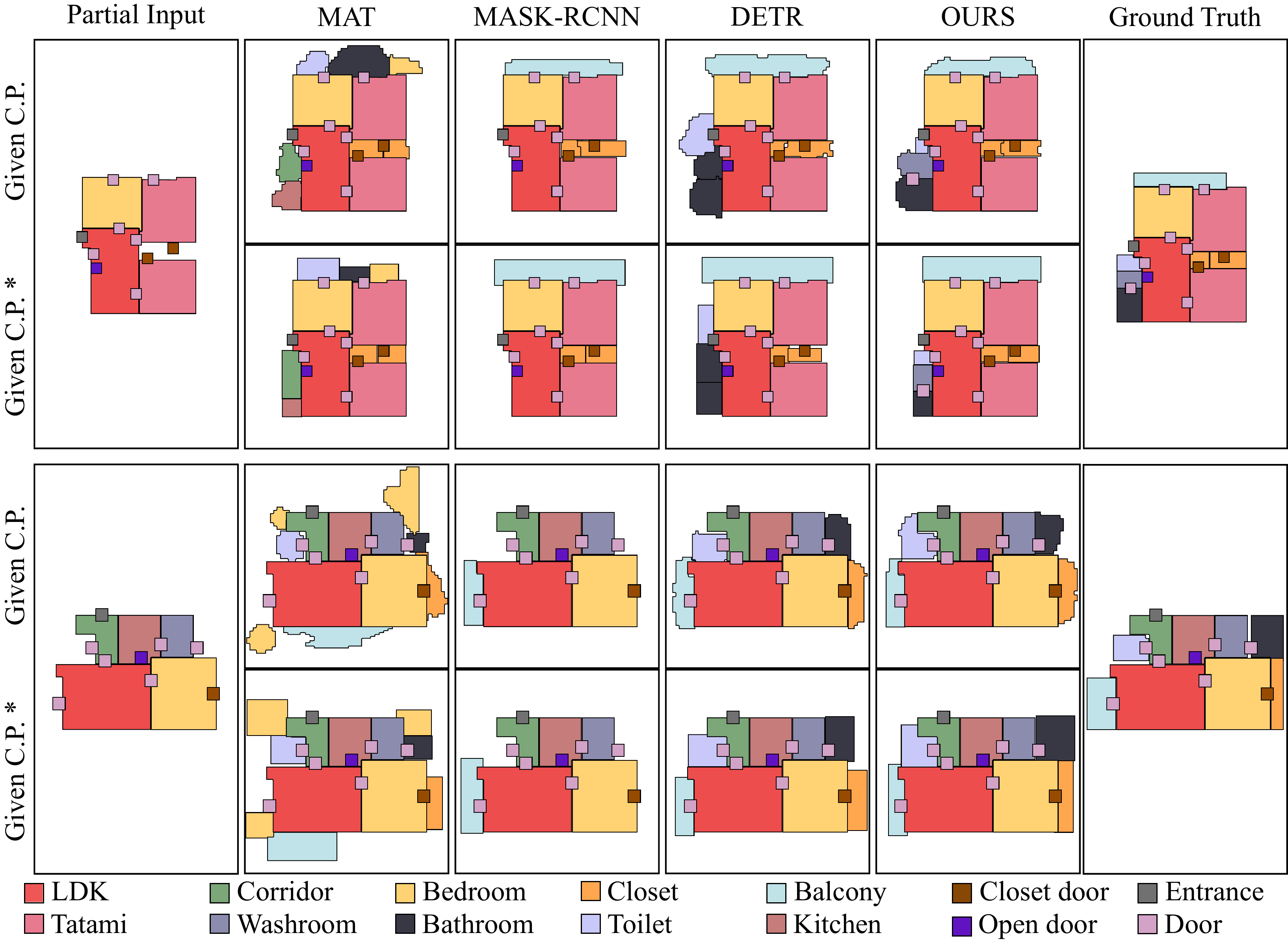}
    \caption{More qualitative results. Reconstructed floorplans before and after the refinement for different baselines and ours by House-GAN++ and several heuristics for more samples. The same refinement process is used for all the methods. The input partial reconstructions are derived either by using given Camera Pose in this figure.}
    \label{fig:sup4}
\end{figure}

 Figure~\ref{fig:sup5} shows reconstructed floorplans by our method before and after applying HouseGAN++.
and~\ref{fig:sup4} present additional output layout samples by our system and other baselines before and after applying refinements. In first row we don't use any refinement, while second row per sample present first and third rows output after applying refinements and HouseGAN++.

\bibliography{egbib}